\documentclass[journal=jctcce, manuscript=article, layout=twocolumn]{achemso}
\usepackage{natbib}
\setkeys{acs}{maxauthors=10, etalmode=truncate}
\usepackage[version=3]{mhchem} 
\usepackage{pgfplots}
\usepackage{tikz}
\usepgfplotslibrary{groupplots}
\pgfplotsset{compat=1.17}
\usepackage{footmisc}

\author{Raul P. Pelaez\footnote{\normalsize{R.P.P. and G.S. contributed equally to this work.}\label{equalcontrib}} }
\author{Guillem Simeon\footref{equalcontrib} }
\author{Raimondas Galvelis}
\affiliation[upf]
{Computational Science Laboratory, Universitat Pompeu Fabra, Barcelona Biomedical Research Park (PRBB), C Dr. Aiguader 88, 08003 Barcelona, Spain.}
\alsoaffiliation[acellera]{Acellera Labs, C Dr Trueta 183, 08005, Barcelona, Spain}
\author{Antonio Mirarchi}
\affiliation[upf]
{Computational Science Laboratory, Universitat Pompeu Fabra, Barcelona Biomedical Research Park (PRBB), C Dr. Aiguader 88, 08003 Barcelona, Spain.}
\author{Peter Eastman}
\affiliation[stanford]
{Department of Chemistry, Stanford University, Stanford, CA 94305, USA}
\author{Stefan Doerr}
\affiliation[acellera]{Acellera Labs, C Dr Trueta 183, 08005, Barcelona, Spain}
\author{Philipp Thölke}
\affiliation[acellera]{Acellera Labs, C Dr Trueta 183, 08005, Barcelona, Spain}
\author{Thomas E. Markland}
\affiliation[stanford]
{Department of Chemistry, Stanford University, Stanford, CA 94305, USA}
\author{Gianni De Fabritiis}
\affiliation[upf]
{Computational Science Laboratory, Universitat Pompeu Fabra, Barcelona Biomedical Research Park (PRBB), C Dr. Aiguader 88, 08003 Barcelona, Spain.}
\alsoaffiliation[acellera]{Acellera Labs, C Dr Trueta 183, 08005, Barcelona, Spain}
\alsoaffiliation[icrea]{Instituci\'o Catalana de Recerca i Estudis Avan\c{c}ats (ICREA), Passeig Lluis Companys 23, 08010 Barcelona, Spain}
\email{g.defabritiis@gmail.com}

\usepackage[utf8]{inputenc}
\usepackage[T1]{fontenc}
\usepackage{float}\tolerance=1000
\usepackage{bm}
\usepackage{svg}
\usepackage{amsmath}
\usepackage{graphicx}
\usepackage{float}
\usepackage{amsmath}
\usepackage{amssymb}
\usepackage{hyperref}
\usepackage{color}
\usepackage{enumerate}
\usepackage{listings}
\SectionNumbersOn
\renewcommand{\vec}[1]{\bm{#1}}

\hypersetup{pdfborder=0 0 0}
\date{\today}
\hypersetup{
 pdftitle={TorchMD-Net v2},
 pdfkeywords={},
 pdfsubject={},
 pdflang={English}}

\title{TorchMD-Net 2.0:  Fast Neural Network Potentials for Molecular Simulations}

\begin{document}
\maketitle

\begin{abstract}
 Achieving a balance between computational speed, prediction accuracy, and universal applicability in molecular simulations has been a persistent challenge. This paper presents substantial advancements in the TorchMD-Net software, a pivotal step forward in the shift from conventional force fields to neural network-based potentials. The evolution of TorchMD-Net into a more comprehensive and versatile framework is highlighted, incorporating cutting-edge architectures such as TensorNet. This transformation is achieved through a modular design approach, encouraging customized applications within the scientific community. The most notable enhancement is a significant improvement in computational efficiency, achieving a very remarkable acceleration in the computation of energy and forces for TensorNet models, with performance gains ranging from 2x to 10x over previous, non-optimized, iterations. Other enhancements include highly optimized neighbor search algorithms that support periodic boundary conditions and smooth integration with existing molecular dynamics frameworks. Additionally, the updated version introduces the capability to integrate physical priors, further enriching its application spectrum and utility in research. The software is available at \url{https://github.com/torchmd/torchmd-net}.

\end{abstract}
\section{Introduction}
\label{sec:intro}

Neural Network Potentials (NNPs)\cite{behlerparrinello, nnp1,nnp2,nnp3,nnp4,nnp5,nnp6} are emerging as a key approach in molecular simulations, striving to optimize the balance between computational efficiency, predictive accuracy, and generality. 

Some software frameworks to facilitate the use of neural network potentials have been developed, such as SchNetPack\cite{schnetpack}, TorchANI\cite{torchani}, DeePMD-Kit\cite{deepmd}, and others. Among the first to appear, we released  TorchMD-Net, initially designed for the Equivariant Transformer architecture\cite{et} and a simpler invariant graph neural network tailored for neural network potentials for protein coarse-graining\cite{majewski2022machine}.
Over time, TorchMD-Net has expanded its model architectures to include TensorNet\cite{simeon2023tensornet}, an $O(3)$-equivariant message-passing neural network utilizing rank-2 Cartesian tensor representations which achieved state-of-the-art accuracy on benchmark datasets. The evolution motivated by the progressive incorporation of different architectures and the changes in the framework needed to accommodate these, positions TorchMD-Net not just as a standalone tool, but as a versatile library for the development of NNPs.

Efficiency has been at the forefront of recent enhancements to TorchMD-Net. Among the optimizations, CUDA graphs have been integrated, providing a performance boost, especially for smaller workloads. TorchMD-Net has also incorporated the latest versions of its key dependencies (mainly PyTorch\cite{pytorch} and PyTorch Lightning\cite{lightning}), with a notable addition being the search for compatibility with the \texttt{torch.compile} submodule from PyTorch 2.0, a feature that compiles Just-In-Time (JIT) modules into optimized kernels. While TorchMD-Net has introduced low precision modes (i.e. \texttt{bfloat16}) primarily as an exploratory tool for researchers, high precision (\texttt{float64}) is also available for ensuring detailed correctness checks during prototyping.

The new technical enhancements include the introduction of periodic boundary conditions, a CUDA-optimized neighbor list, and memory-efficient dataset loaders. The inclusion of TorchMD-Net in the conda-forge\cite{conda_forge} package repository and the release of the documentation \cite{torchmdnetdocs} are steps taken to enhance its accessibility to researchers.
Another feature is TorchMD-Net's capacity to blend empirical physical knowledge into NNPs via priors. The integration of atom-wise and molecule-wise priors, such as the Ziegler-Biersack-Littmark \cite{zblprior} and Coulomb potentials, allows for a more nuanced approach in simulations.

TorchMD-Net emphasizes compatibility with leading molecular dynamics (MD) packages, especially with OpenMM\cite{openmm8}. OpenMM, widely recognized in the computational chemistry field, can now interface directly with TorchMD-Net through the OpenMM-Torch\cite{openmmtorch} plugin. This integration has been a collaborative effort, with OpenMM-Torch being co-developed by the core teams of both OpenMM and TorchMD-Net. This ensures streamlined and effective utilization of TorchMD-Net models within OpenMM's simulation framework.

In the following sections we provide an overview of the TorchMD-Net framework. The manuscript is ordered as follows. In the Methods section, we initially provide a schematic overview of the principal model components, to then go over the currently available NNP architectures. We continue in subsection Training with details about the different parts involved in the training and deployment of these architectures and how they are exposed in TorchMD-Net. Then, in the Optimization subsection, we lay out the optimization strategies employed in this release. Finally, we present a series of validation and performance results in the Results section.

TorchMD-Net is freely available with a permissive license (MIT) at \url{https://github.com/torchmd/torchmd-net}.

\section{Background}\label{sec:methods}
We interpret a  neural network potential as a machine learning model that takes as input a series of atomic positions, denoted by $\vec{R}$, embedding indices such as atomic numbers, $Z$, and optionally charges (which might be per-sample or per-atom)\cite{qs}, $q$, and outputs a per-sample scalar value and optionally its negative gradient with respect to the positions, typically interpreted as the potential energy and atomic forces, respectively. Note, however, that TorchMD-Net is  not limited to this interpretation of the outputs, which are generally labeled as \texttt{y} and \texttt{neg\_dy} respectively. 

\begin{figure*}
    \centering
    \includegraphics[width=\linewidth]{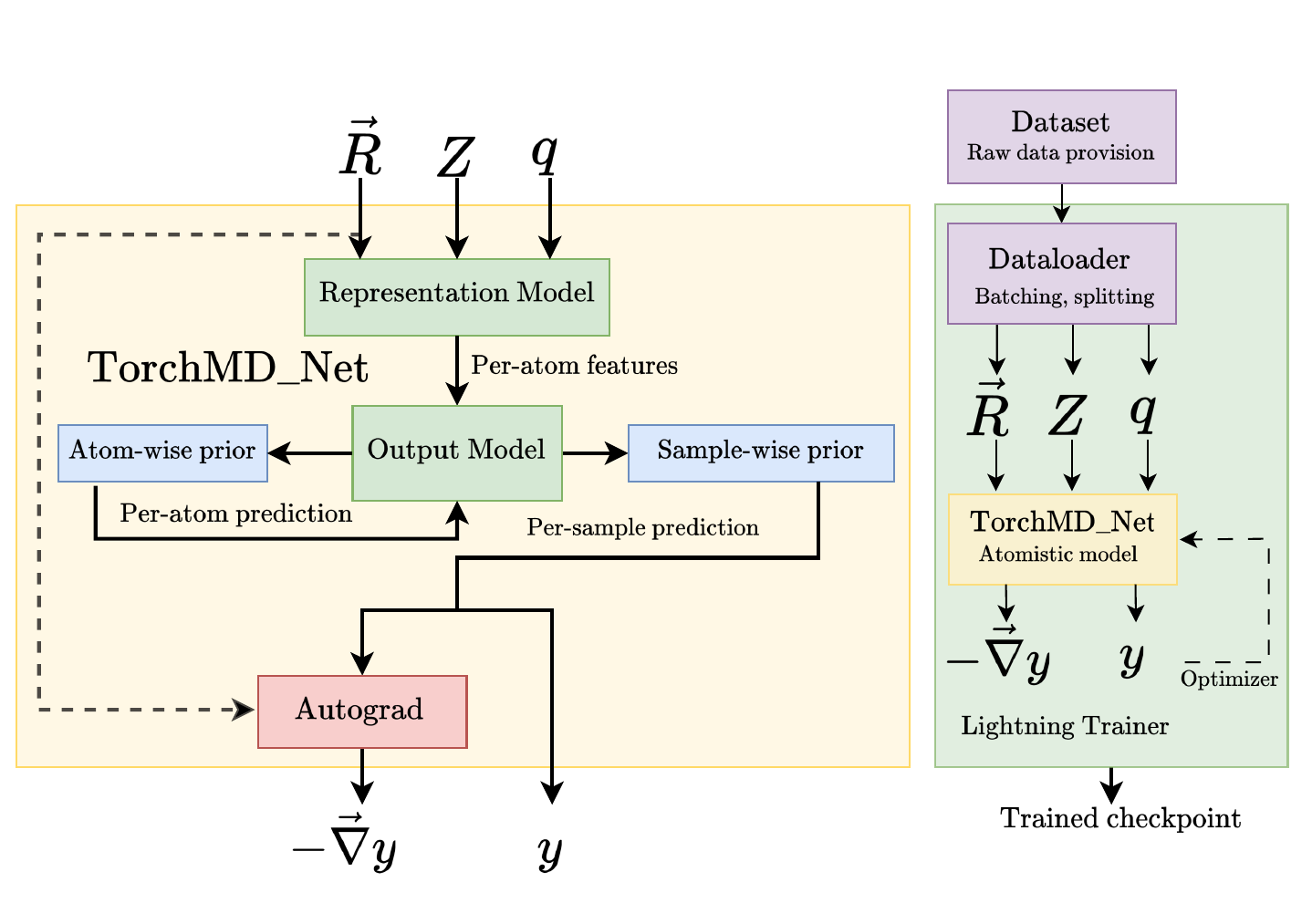}
    \caption{The main module in TorchMD-Net is called \texttt{TorchMD\_Net} from the \texttt{torchmdnet.models.model} module. This class combines a  given representation  model
    (such as  the Equivariant Transformer),  an output model  (such as
    the scalar output  module) and a prior model (such  as the Atomref
    prior), producing a  module that takes as input a  series of atoms
    features and  outputs  a scalar value  (i.e.  energy  per molecule) and when \texttt{derivative = True}, its negative gradient with respect to the input positions (i.e. atomic forces).}
    \label{fig:tmdnet_class}
\end{figure*}

Figure \ref{fig:tmdnet_class} provides a comprehensive overview of the TorchMD-Net architecture. The diagram's left section illustrates the various components of the primary module, designated as \texttt{TorchMD\_Net}, which constitutes, conceptually and in the API itself, an NNP model. Each component within this object is modular and customizable, allowing for the creation of diverse models. At the heart of the NNP is the \texttt{representation\_model}. This part of the architecture takes the set of inputs stated above and outputs a series of per-atom features. These features are subsequently fed into an \texttt{output\_model}. The purpose of this model is to further process these features into single atomic values, which typically will be aggregated and will represent the total potential energy, though it can represent other per-sample or per-atom quantities as well, depending on the specifics of its design and (optional) aggregation scheme. Output models normally include learnable parameters (e.g. a multilayer perceptron).
Prior physical models can be employed to augment either the atom-level or aggregated per-molecule predictions with further physical insights. Furthermore, the framework integrates PyTorch's Autograd for automatic differentiation, enabling the computation of the negative gradient of the per-molecule scalar prediction with respect to atomic positions. This is particularly relevant when interpreting the per-molecule value as the potential energy, as it yields the atomic forces in a way that ensures, by construction, that the resulting force field is energy-conserving.

This modular logic allows for flexibility in the combination of representation models and output models. Therefore, by building a custom output module, researchers can make use of the representation models for other prediction tasks beyond potential energy and forces. 

\subsection{Available representation models}

Although the framework is not restricted to it, current models in TorchMD-Net are message-passing neuralnetworks\cite{messagepassing1,messagepassing2} (MPNNs) which learn approximations to the many-body potential energy function. Atoms are identified with graph nodes embedded in 3D space, building edges between them after the definition of some cutoff radius. The neural network uses atomic and geometric information to learn expressive representations by propagating, aggregating, and transforming features from neighboring nodes found within the cutoff radius\cite{expressive, hitchhiker}. In most current NNPs, after several message-passing steps, node features are used to predict per-atom scalar quantities which are identified with atomic contributions to the energy of the molecule.

\subsubsection{TensorNet}

TensorNet\cite{simeon2023tensornet} is an $O(3)$-equivariant model based on rank-2 Cartesian tensor representations. Euclidean neural network potentials\cite{e3nn, nequip, allegro, mace} have been shown to achieve state-of-the-art performance and better data efficiency than previous models, relying on higher-rank equivariant features which are irreducible representations of the rotation group, in the form of spherical tensors. However, the computation of tensor products in these models can be computationally demanding. In contrast, TensorNet exploits the use of Cartesian rank-2 tensors (3x3 matrices) which can be very efficiently decomposed into scalar, vector and rank-2 tensor features. Furthermore, Clebsch-Gordan tensor products are substituted by straightforward and node-level 3x3 matrix products. 

TensorNet achieved state-of-the-art accuracy on common benchmark datasets with a small number of message-passing layers, learnable parameters, and computational cost. The prediction of up to rank-2 molecular properties that behave appropriately under geometric transformations such as reflections and rotations is also possible.

\subsubsection{Equivariant Transformer}

The Equivariant Transformer\cite{et} (ET) is an equivariant neural network that uses both scalar and Cartesian vector representations. The distinctive feature of the ET in comparison to other Cartesian vector models such as PaiNN\cite{painn} or EGNN\cite{egnn} is the use of a distance-dependent dot product attention mechanism, which achieved comparable performance to state-of-the-art models on benchmark datasets at the time of publication.Furthermore, the analysis of attention weights allowed us to extract insights into the interaction of different atomic species for the prediction of molecular energies and forces. The model also exhibits a low computational cost for inference and training in comparison to some of the most used NNPs in the literature\cite{egraffbench}. 

As part of the current release, we removed a discontinuity at the cutoff radius. In the original description, vector features' residual updates, as opposed to scalar features' updates, received contributions from the value pathway of the attention mechanism which were not properly weighted by the cosine cutoff function envelope, which is reflected in Eq. 9 in the original paper\cite{et}. We fixed it by applying $\phi(d_{ij})$, i.e., $\mathrm{split}(V_j \odot D^{V}_{ij}) \rightarrow \mathrm{split}(\phi(d_{ij})  V_j \odot D^{V}_{ij})$. To ensure backward compatibility, this modification is only applied when setting the new ET argument \texttt{vector\_cutoff = True}. The impact of this modification is evaluated in the results section.

\subsubsection{Graph Network}

The graph network is an invariant model inspired by both the SchNet\cite{Schutt2017} and PhysNet\cite{Unke2019-tb} architectures. Described in Ref~\citenum{majewski2022machine}, the network was optimized to have satisfactory performance on coarse-grained proteins, allowing the building of NNPs that correctly reproduce fast-folder protein free energy landscapes. In contrast to the ET and TensorNet, the graph network only uses relative distances between atoms as geometrical information, which are invariant to translations, rotations, and reflections. The distances are used by the model to learn a set of continuous filters that are applied to feature graph convolutions as in SchNet\cite{Schutt2017}, progressively updating the initial atomic embeddings by means of residual connections.

\subsection{Physical priors for models}
Priors are additional physical terms that can be introduced for the prediction of potential energies. Some of these terms have been used in NNPs in the literature \cite{physnet, spookynet,majewski2022machine}, sometimes even including learnable parameters. In TorchMD-Net, we provide some predefined priors, which can be optionally added to the neural network energy prediction as additional physics-based contributions: 
\begin{itemize}
    \item Atomref: These are per-element atomic reference energies, which are usually provided directly in the dataset. In this case, the neural network has to predict the remaining contribution to the potential energy, which can be regarded as the formation energy of the molecule. There is also the option of making this prior learnable, in which case it is initialized with atomic reference energies, but these contributions are modified during training.
    \item Coulomb: This prior corresponds to the usual Coulomb electrostatic interaction, scaled by a cosine switching function to reduce its effect at short distances. Using this prior requires providing per-atom partial charges.
    \item D2 dispersion: In this case, the prior corresponds to the D2 dispersive correction used in DFT-D2 \cite{d2prior}. $C_{6}$ coefficients and Van der Waals radii for elements are already incorporated in the method.
    \item ZBL potential: This prior implements the Ziegler-Biersack-Littmark (ZBL) potential for screened nuclear repulsion as described in Ref~\citenum{zblprior}. It is an empirical potential effectively describing the
    repulsion between atoms at very short distances, and only atomic numbers need to be provided. 
\end{itemize}

Note that forces are computed directly by \texttt{autograd} when adding the energy contributions coming from the priors before the backward automatic differentiation step. Even though the previous terms are the currently predefined options in TorchMD-Net, all these priors are derived from a general \texttt{BasePrior} class, which easily allows researchers to implement their own priors, following the modular logic behind the framework.

\subsection{Training}\label{sec:training}

The right diagram in Figure \ref{fig:tmdnet_class} depicts the main training loop in TorchMD-Net. A Dataset provides sample/output pairs for the NNP and is divided into training, validation and testing sets and batched by a Dataloader (as provided by the Pytorch Geometric library\cite{geometric}). We make use of the PyTorch Lightning library's\cite{lightning} trainer, which also allows multi-GPU training. Checkpoints are generated during training, containing the current weights of the model, which can then be subsequently loaded for inference or further training.

\subsubsection{Datasets} \label{sec:datasets}

Within TorchMD-Net, datasets can be accessed through the YAML configuration file for use with the \texttt{torchmd-train} utility or programmatically via the Python API. 
Predefined datasets include SPICE\cite{spice}, QM9\cite{qm9}, WaterBox~\cite{water}, (r)MD17\cite{Chmiela}\cite{rmd17}, MD22\cite{md22}, ANI1\cite{ani1}, ANI1x\cite{ani1x1}, ANI1ccx\cite{ani1x1}, ANI2x\cite{ani2x} and the COMP6\cite{ani1x2} evaluation dataset with all its subsets (ANIMD, DrugBank, GDB07to09, GDB10to13, Tripeptides and S66X8), offering diverse training environments for molecular dynamics and quantum chemistry applications. These datasets serve as common benchmarks in the field of neural network potentials. However, on top of these, the framework allows the flexible incorporation of user-generated datasets for customized applications. The \texttt{Custom} dataset functionality allows users to train models with molecular data encapsulated in simple NumPy file formats without writing a single line of code. By specifying paths to coordinate and embedding index (e.g. atomic numbers) and reference energy and force files, researchers can easily integrate their datasets into the training process. This capability ensures TorchMD-Net's adaptability to a wider array of applications beyond its pre-packaged offerings. In addition, TorchMD-Net offers support for other popular dataset formats, such as HDF5.
Special care is taken to ensure data is cached as much as possible, using techniques such as in memory datasets and memory mapped files.

\subsubsection{Losses}
During training, a weighted sum of mean squared error (MSE) losses of energy and forces is used, weighting each of them according to user input. In validation, we provide both L1 and MSE losses separately for energies and forces, while for testing L1 losses alone are used. The framework allows to use of an exponential moving average (EMA) to update the losses during the training and validation stages to smooth out the progression of loss values.

\subsection{Usage examples}
In the following sections we showcase code for some typical use cases of TorchMD-Net. While these snippets are generally self-contained the reader is pointed to the online documentation \cite{torchmdnetdocs} for further information.
\subsubsection{Training code example}
The project introduces a command line tool, \texttt{torchmd-train}, designed as a code-free method for model training. This tool is set up through a YAML file, with several examples available in the TorchMD-Net GitHub repository for reference. However, we also offer an illustrative script here that outlines the process of training an existing model using the Python API. The \texttt{LNNP} class, found within the \texttt{torchmdnet.module} module, encapsulates the procedures for both the creation and training of a model. This class is inherited from Pytorch Lightning \texttt{LightningModule}, offering all the extensive customization available in it. The following is a succinct yet comprehensive example of how to utilize \texttt{LNNP} for training purposes:
\begin{lstlisting}[language=Python]
from torchmdnet.data import DataModule
from torchmdnet.module import LNNP
from pytorch_lightning import Trainer
args = {
    'dataset': 'ANI1X',
    'model': 'tensornet',
    'num_epochs': 200,
    'embedding_dimension': 128,
    'num_layers': 2,
    'num_rbf': 32,
    'rbf_type': 'expnorm',
    'trainable_rbf': False,
    'activation': 'silu',
    'cutoff_lower': 0.0,
    'cutoff_upper': 5.0,
    'max_z': 100,
    'max_num_neighbors': 64,
    'derivative': True # So the model returns forces.
}
data = DataModule(args)
data.prepare_data()
data.setup("fit")
lnnp = LNNP(args, 
prior_model=None, 
mean=data.mean, 
std=data.std)
trainer = Trainer(max_epochs=args['num_epochs'])
trainer.fit(lnnp, data)
model = LNNP.load_from_checkpoint(trainer.checkpoint_callback.best_model_path)
trainer = pl.Trainer(inference_mode=False)
trainer.test(model, data)
\end{lstlisting}
This example shows the minimal steps required to prepare data, initialize the \texttt{LNNP} class, train and test a model using PyTorch Lightning's \texttt{Trainer}. The \texttt{Trainer} here is simplified for brevity; in practice, additional callbacks and logger configurations could be added.

\subsubsection{Loading a Trained Model for Inference}

After training a model, the next logical step is to use it for inference. TorchMD-Net offers a dedicated function, \texttt{load$\_$model}, to facilitate this. Below is a concise example:

\begin{lstlisting}[language=Python]
from torchmdnet.models.model import load_model
# Define the path to the saved model checkpoint
checkpoint_path = "path/to/saved_model_checkpoint.ckpt"
# Load the model
loaded_model = load_model(checkpoint_path)
# Prepare the input data (atomic numbers, positions, batch index, etc.)
# For demonstration, these are placeholders and should be replaced with actual data
input_data = {
    'z': torch.Tensor([...]),
    'pos': torch.Tensor([...]),
    'batch': torch.Tensor([...])
    # ... other optional fields
}
# Perform inference
energy, forces = loaded_model(**input_data)
# Energy and forces are now available for further analysis or visualization
\end{lstlisting}

In this example, \texttt{checkpoint$\_$path} should point to the location where the trained model checkpoint is saved. The \texttt{input$\_$data} dictionary should be populated with the actual atomic numbers, positions, and other required or optional fields. Finally, \texttt{energy} and \texttt{forces} are obtained from the loaded model and can be used as needed.

\subsubsection{Integration with OpenMM}

It is possible to run TorchMD-Net neural network potentials as force fields in OpenMM\cite{openmm8} to run molecular dynamics. The OpenMM-Torch\cite{openmmtorch} package is leveraged for this. Integration consists of writing a wrapper class to accommodate the unit requirements of OpenMM and to provide the model with any information not proper to OpenMM (like the embedding indices). The following code showcases an example of how to add a TorchMD-Net NNP as an OpenMM Force.

\begin{lstlisting}[language=Python]
from torchmdnet.models.model import load_model
from openmmtorch import TorchForce
from openmm import System

class Wrapper(torch.nn.Module):

    def __init__(self, embeddings, model):
        super(Wrapper, self).__init__()
        self.embeddings = embeddings
        # OpenMM will compute the forces 
        #  by backpropagating the energy,
        # so we can load the model with derivative=False
        self.model = load_model(model, derivative=False)

    def forward(self, positions):
        # OpenMM works with nanometer positions 
        #  and kilojoule per mole energies
        # Depending on the model, you might need
        #  to convert the units
        positions = positions * 10.0 # nm -> A
        energy = self.model(z=self.embeddings, pos=positions)[0]
        return energy * 96.4916 # eV -> kJ/mol

# The embeddings used during training (e.g. atomic numbers)
#  for each atom in the simulation.
z = torch.tensor([1,1], torch.long) 
model = torch.jit.script(Wrapper(z, "model.ckpt"))
# Create a TorchForce object from the model
torch_force = openmmtorch.TorchForce(model)
system = System()
# The TorchForce object can be used as a regular OpenMM Force
system.addForce(torch_force)
# Set up the rest of the OpenMM simulation
# ...
\end{lstlisting}

\subsection{Optimization techniques}\label{sec:opt}
Typical neural network potential (NNP) algorithms implemented in PyTorch\cite{pytorch} comprise a series of sequential operations such as multilayer perceptrons and message-passing operations. 

As PyTorch operations translate into highly optimized CUDA kernel calls, and owning to the eager-first nature of PyTorch, the efficiency of modern GPUs often turns kernel launching overhead into a performance bottleneck. CUDA graphs address this by consolidating multiple kernel calls into a single graph, drastically reducing kernel launch overhead. However, CUDA graphs impose stringent limitations. These include the need for static shapes in graphed code sections, which can lead to costly recompilations or memory inefficiencies, and the exclusion of operations requiring CPU-GPU synchronization.

Conversely, developments in the compiler community\cite{dlcompiler} , including technologies like OpenAI's Triton\cite{triton} and subsequently PyTorch enhancements, are gradually diminishing the reliance on CUDA graphs by automatically changing the structure of the code in ever more profound ways (i.e kernel fusion\cite{fusionRNN,fusion2010,fusionblas}). These advancements, such as TorchDynamo introduced in PyTorch 2.0 through \texttt{torch.compile}, optimize code structure through Just-In-Time (JIT) compilation. 

Even with JIT, and in general transpilation-based techniques, CUDA graphs often provide the best out-of-the-box performance improvements and at the bare minimum, facilitate the optimizations introduced by the former.
Encapsulating a piece of code within a CUDA graph, a process known as 'stream capture', necessitates adherence to several specific requirements. This often demands substantial modifications to the code. Crucially, for code to be eligible for capture, it must avoid any CPU-GPU synchronization activities, including synchronization barriers and memory copies. Additionally, all arrays involved in the operations must possess static shapes and fixed memory addresses, precluding any dynamic memory allocations during the process.

The CUDA graph interface in PyTorch alleviates many challenges associated with adapting code for stream capture. It particularly excels in managing memory allocations within captured environments automatically and transparently. However, challenges arise in specific implementations, as exemplified by TensorNet. The main issue in TensorNet is its neighbor list, which inherently varies in shape at each inference step due to the fluctuating number of neighbors. This variation affects the early stages of the architecture, resulting in TensorNet primarily operating on dynamically shaped tensors. To address this, we implemented a static shape mode that creates placeholder neighbor pairs up to a predetermined maximum. We then ensure the architecture disregards these placeholders' contributions. Although this method increases the initial workload, our empirical data indicates that the performance gains from capturing the entire network substantially outweigh this added overhead.
Work on static shaped graph networks has been done before in JAX libraries such as jgraph\cite{jraph2020github}.

Regardless of the choice of framework, the aforementioned general optimization techniques (and in particular complying with CUDA graph requirements) constitute a good roadmap in the quest for efficiency. Many of these frameworks are down the line aiming to somehow transform user code into a series of CUDA (or some other massively parallel language with similar properties such as Triton or OpenCL) kernels, with a preferably small number of them. General strategies like eliminating CPU-GPU synchronization barriers or communication help in this regard. The generality of these directives makes the resulting implementations more aligned with \texttt{torch.compile} in our case, but would for instance also aid \texttt{jax.jit} if a developer were to port our neighbor list to JAX.

By making our operators strive for CUDA-graph compatibility, making them \texttt{torch.compile}-aware and easy to import, we intend to boost the development of future architectures and provide efficient tools for the community even outside the TorchMD-Net ecosystem.

In the following sections, we explore the impact of these optimizations on both inference and training performance.

\subsubsection{Neighbor search and periodic boundary conditions}

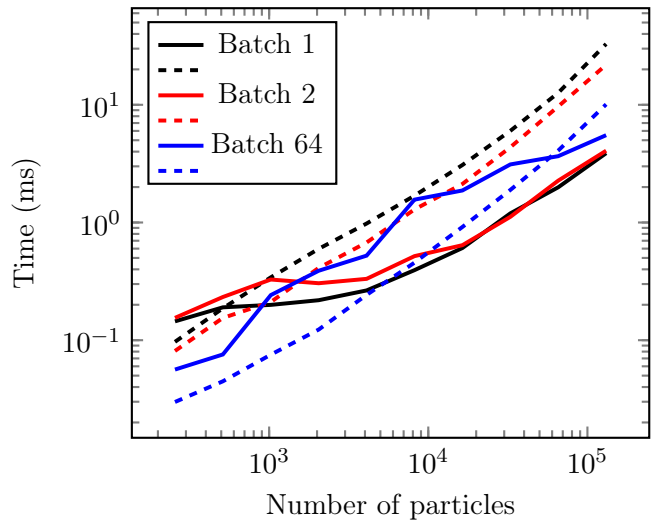
\begin{figure}
\begin{tikzpicture}
  \newcommand{\thickness}{1.5pt}
\begin{axis}[
    xmode=log, ymode=log,
    width=\linewidth,
    cycle list={
      {black, line width=\thickness},
      {black, dashed, line width=\thickness},
      {red, line width=\thickness},
      {red, dashed, line width=\thickness},
      {blue, line width=\thickness},
      {blue, dashed, line width=\thickness},
    },
    axis line style={thick},
    tick style={thick},
    legend pos = north west,
    legend style={draw, thick, font=\small},
    label style={font=\small},
    tick label style={font=\small},
    xlabel={Number of particles},
    ylabel={Time (ms)},
    title={},
    ]
\addplot table [x=particles, y=cell, col sep=comma] {neighbor_benchmark/results_particles_1.csv};
\addplot table [x=particles, y=shared, col sep=comma, dashed] {neighbor_benchmark/results_particles_1.csv};
\addlegendentry{Batch 1}
\addlegendentry{}
\addplot table [x=particles, y=cell, col sep=comma] {neighbor_benchmark/results_particles_2.csv};
\addplot table [x=particles, y=shared, col sep=comma, dashed] {neighbor_benchmark/results_particles_2.csv};
\addlegendentry{Batch 2}
\addlegendentry{}
\addplot table [x=particles, y=cell, col sep=comma] {neighbor_benchmark/results_particles_64.csv};
\addplot table [x=particles, y=shared, col sep=comma, dashed] {neighbor_benchmark/results_particles_64.csv};
\addlegendentry{Batch 64}
\addlegendentry{}
\end{axis}
\end{tikzpicture}
\caption{Performance comparison of cell (solid line) and brute-force (dashed line) neighbor search strategies across different batch sizes for a random cloud of particles with $64$ neighbors per particle on average. Cell list performance tends to degrade with increasing batch size, while the opposite is true for brute force.}
\label{fig:batch}
\end{figure}

\begin{figure}
\begin{tikzpicture}
  \newcommand{\thickness}{1.5pt}
\begin{axis}[
    xmode=log,
    log basis x={2},
    log ticks with fixed point,
    xmax=1024,
    width=\linewidth,
    cycle list={
      {black, line width=\thickness},
      {black, dashed, line width=\thickness},
    },
    axis line style={thick},
    tick style={thick},
    legend style={draw, thick, font=\small},
    label style={font=\small},
    tick label style={font=\small},
    xlabel={Batch Size},
    ylabel={Time (ms)},
    title={},
    ]
\addplot table [x=batch, y=cell, col sep=comma] {neighbor_benchmark/results_batch.csv};
\addlegendentry{Cell}
\addplot table [x=batch, y=shared, col sep=comma, dashed] {neighbor_benchmark/results_batch.csv};
\addlegendentry{Brute force}
\end{axis}
\end{tikzpicture}
\caption{Performance comparison of cell (solid line) and brute-force (dashed line) neighbor search strategies across different batch sizes for a random cloud of $32$k particles with $64$ neighbors per particle on average. The particles are split into a certain number of batches.}
\label{fig:batch1}
\end{figure}
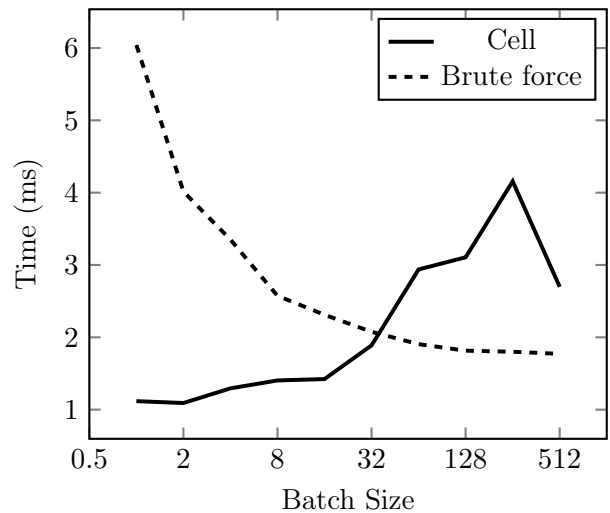

Message-passing neural networks, such as the architectures currently supported in the framework, require a list of connections among nodes referred to as edges. This list is constructed by proximity after the definition of a cutoff radius (a neighbor list). TorchMD-Net offers a neighbor list construction engine specifically tailored for NNPs, exposing a naive brute-force $O(N^2)$ algorithm that works best for small workloads and a cell list (a standard $O(N)$ hash-and-sort strategy widely used in MD\cite{celllist, celllist2}) that performs better for large systems (see Figures \ref{fig:batch} and \ref{fig:batch1}). Effectively, this engine makes neighbor search a negligible part of the overall computation. This operation is exposed as a PyTorch Autograd extension with a reference pure PyTorch CPU implementation. The forward pass has a registration for inputs in a CUDA device written in C++/CUDA. Finally, a custom implementation of the backward pass is also included and written in PyTorch to easily accommodate for higher order derivatives\footnote{The second backward pass of this operation is needed when training using forces in TorchMD-Net, as the forces are computed via backpropagation of the energy.}. From a user perspective this translates into the operation being usable in an Autograd environment from any device compatible with PyTorch.

Special measures are taken into account to ensure that the neighbor search is compatible with CUDA-graphs. For this matter, it is required that the neighbor search works on a statically-shaped set of input/outputs, which poses a problem given that the number of neighbor pairs in the system is not known in advance and is bound to change from input to input. We solve this by requiring an upper bound for the number of pairs in the system and padding the outputs with a special value ($-1$) for unused pairs. Furthermore, TorchMD-Net architectures support rectangular and triclinic periodic boundary conditions.

Contrary to usual MD workloads, it is common to have batches of input samples in NNPs. This owns to the very nature of neural network training but also can benefit inference (for instance, allowing the possibility of running many simulations in parallel, like TorchMD\cite{torchmd} does). Our neighbor list is able to handle arbitrary batch sizes while maintaining compatibility with CUDA graphs.
There are many implementations of various neighbor list algorithms in external packages but, regardless of their performance, none of them fully satisfy our set of constraints. These being that it should be available in pytorch (discarding, for instance, any JAX-centered library such as JAX-MD\cite{jaxmd2020}), support CUDA-graphs (discarding the \texttt{radius\_graph} function in Pytorch-Geometric\cite{geometric}), be able to handle batches (discarding ASE\cite{ase-paper} and PyMatGen\cite{pymatgen}) and finally support rectangular and triclinic periodic boundary conditions (which some of the aforementioned implementations lack).

The current cell list implementation constructs a single cell list including atoms for all batches, excluding pairs of particles that pertain to different batches when finding neighbors in it. This makes it so that each particle has to perform a check against every other particle in the vicinity for all batches, which degrades performance with increasing batch size. We find this to be an acceptable compromise given that doing it this way facilitates compatibility with CUDA graphs and we assume that with an increasing number of particles (where the cell list excels) the typical batch size will decrease.
Still, the particularities of the cell list implementation make its performance especially susceptible to the batch size, as evidenced by the variability observed in the cell list curves in figures \ref{fig:batch} and \ref{fig:batch1}.

Our approach to handling batches consists of assuming there is only one batch, delaying this check-up until just before fetching the atom positions to check their distances. While the scalability of this approach with the number of batches is questionable (many unnecessary pairs might be checked) it allows us to overcome a series of limitations besides the aforementioned for the cell list. The operation takes as principal inputs a contiguous tensor of atomic positions and another with batch indices (indicating to which batch an atom pertains). Splitting these tensors as a precomputation would in general require knowing the number of distinct batches (involving a costly reduction operation and synchronization barriers if the value is required CPU-side) as well as computing where each one starts and ends in the positions tensor (assuming the input is sorted by batch). Even if the batch locations were known in advance, calling the neighbor operation for each batch independently could result in the launching of several instances of small CUDA kernels, which would not fill the GPU and hurt performance overall. On the other hand, an approach with a control flow depending on the contents of the input tensors would hinder the static requirements of CUDA graphs.

We offer two different brute-force neighbor search strategies, which are selected depending on the number of atoms received by the operation. In both of them, the neighbor list is generated in a single CUDA kernel in which every possible pair in the system is checked, first to ensure both are in the same batch and then for their distance. The range of applicability of these algorithms (intended for workloads of less than 10 thousand atoms) allows for most, if not all, of the inputs to be read fit in the deepest cache levels of the GPU\footnote{For reference, the RTX4090 has 128KB of L1 cache (the fastest access global memory cache) per streaming multiprocessor, which is theoretically enough to hold the data for the positions of $10^4$ atoms stored in single precision ($10^4\text{(atoms)}3 \text{(coordinates)} 4\text{(bytes/value)} \approx 117KB$) }, leaving these kernels typically bottlenecked by the actual writing of the neighbor pairs, which is carried out using an atomic counter to "reserve" a space in the neighbor list and then writing the pair to this list. The first brute-force strategy launches a thread per possible pair, this limits its function to less than $2^{16}-1$ atoms given that the number of possible pairs in this case, $\frac{N(N-1)}{2}$, coincides with the maximum number of threads that can be launched in a single CUDA kernel, $2^{31}-1$. The other strategy is the Fast N-Body algorithm described in chapter 31 of GPU-Gems 3\cite{celllist}, which launches a thread per atom and leverages CUDA's shared memory. Remarkably, we find the thread-per-pair strategy to be the fastest of the two up until its theoretical limit of $2^{16}-1$ atoms. Figure \ref{fig:batch1} shows how the brute algorithm (in this case the thread-per-pair implementation) scales well with the number of batches, as the loading of the positions can be mostly avoided and the batch tensor fits in the cache in its entirety.

All data presented in this section was gathered in an RTX4090 NVIDIA GPU using CUDA 12. Each point is obtained by averaging 50 identical executions. Warmup executions are also performed before measuring.

\subsubsection{Training}
Optimizing neural network training presents distinct challenges compared to inference optimization. Primarily, the variable length of each training sample, exacerbated by batching processes (where a varying number of samples constitutes a single network input), impedes optimizations dependent on static shapes (i.e. CUDA graphs). A potential solution involves integrating 'ghost molecules', akin to strategies used in static neighbor list shaping, to standardize the atom count inputted to the network. However, this method increases memory consumption in an already memory-constrained environment and can potentially be wasteful when the dataset presents samples highly heterogeneous in size.

Moreover, training necessitates backpropagation through the network. In our context, this involves a double backpropagation process when the loss function includes force calculations. Currently, double backpropagation is inadequately supported by the PyTorch compiler. A workaround is to manually implement the network's initial backward pass (specifically, the force computation). This adjustment enables Autograd to perform only a single backward pass during training, leveraging the PyTorch compiler's capabilities. Nevertheless, challenges persist with the PyTorch compiler when managing dynamic input shapes. We hope future versions of PyTorch will improve support for both dynamic shapes and double backpropagation, allowing to make use of our inference optimizations in training seamlessly.

Given the current constraints, the current release does not include any training-specific optimizations besides the improved dataloader support as previously described.
\section{Results}\label{sec:results}
\subsection{Validation}

In this subsection, we evaluate the impact of the architectural modifications introduced in the models on predictive accuracy. In the case of TensorNet the modifications targeted its computational performance alone, while for the ET one needs to consider the changes induced by \texttt{vector\_cutoff = True}.  

\subsubsection{Accuracy with TensorNet}
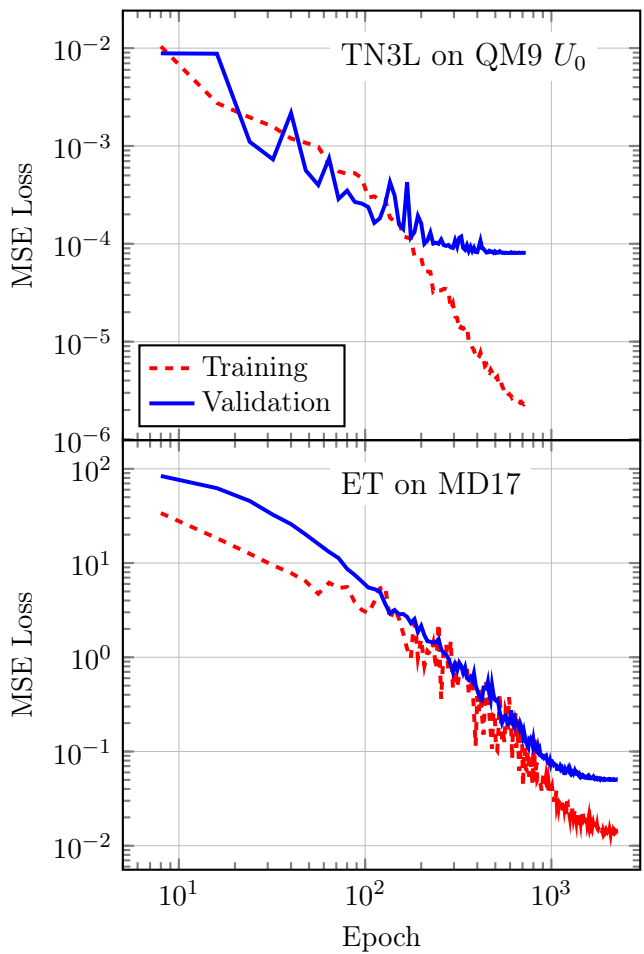
\begin{figure}
\centering
\begin{tikzpicture}
  \newcommand{\thickness}{1.5pt}
  \begin{groupplot}[
    group style={
        group size=1 by 2,
        horizontal sep=0cm,
        vertical sep=0cm,
    },
    title style={at={(0.5,1.2)}},
    xmode=log, ymode=log,
    grid=major,
    legend pos=south west,
    width=1.0\linewidth, 
    cycle list={
      {red,dashed, line width=\thickness},
      {blue, line width=\thickness},
      {green,dashed, line width=\thickness},
      {orange, line width=\thickness},
      {black, line width=\thickness},
    },
    each nth point=8, 
    axis line style={thick},
    tick style={thick},
    legend style={draw, thick, font=\small},
    legend cell align={left},
    label style={font=\small},
    tick label style={font=\small},
    xmin=5,
    xmax=3000
    ]
\nextgroupplot[
        ylabel={MSE Loss},
        xticklabel=\empty
    ]
\addplot table [x=epoch,y=train_y_mse_loss, col sep=comma] {latest_qm9.csv};
\addlegendentry{Training}
\addplot table [x=epoch, y=val_y_mse_loss, col sep=comma] {latest_qm9.csv};
\addlegendentry{Validation}
\node[anchor=north west, fill=white] at (axis description cs:0.4,0.95) {TN3L on QM9 $U_0$};
    \nextgroupplot[
        xlabel={Epoch},
        ylabel={MSE Loss}
    ]
\addplot table [x=epoch, y=train_y_mse_loss, col sep=comma] {fixed_et_md17.csv};
\addplot table [x=epoch, y=val_y_mse_loss, col sep=comma] {fixed_et_md17.csv};
\node[anchor=north west, fill=white] at (axis description cs:0.4,0.95) {ET on MD17};
  \end{groupplot}
\end{tikzpicture}
\caption{Training and validation curves for two different training benchmark datasets. Up: TensorNet with 3 interaction layers on the QM9 $U_0$, parameters in table S3.
Down: Equivariant Transformer on MD17, parameters in Table S2.
}
\label{fig:trainings}
\end{figure}

Original test MAE presented in Ref.~\citenum{simeon2023tensornet} for the QM9 $U_0$ target quantity is $3.9(1)$ meV, while the latest optimized versions of the model (see Figure \ref{fig:trainings}) yield $3.8(2)$ meV, confirming that the architectural optimizations do not affect TensorNet's prediction performance. The training loss was computed in this case as the MSE between predicted and true energies. This state-of-the-art performance is achieved with the largest model with 4.0 million trainable parameters, with specific architectural and training hyperparameters being found in Table S3.
We also provide in Table \ref{tab:qm9r} the accuracy of smaller and shallower models on the same QM9 quantity (that is, using the same hyperparameters as in Table S3,
except for \texttt{embedding\_dimension = 128} and \texttt{num\_layers = 0, 1, 2}), while comparing them to other NNPs. Overall, TensorNet demonstrates very satisfactory performances, achieving close to state-of-the-art accuracy ($<5$ meV MAE) with a very reduced number of parameters.


\subsubsection{Accuracy with the Equivariant Transformer}

As previously mentioned, we provide an implementation of the ET where it is modified by applying the cutoff function to the values' pathway of the attention mechanism to enforce a continuous energy landscape at the cutoff distance. Therefore, we checked to which extent these changes, together with TorchMD-Net's ones, affect the accuracy of the Equivariant Transformer. 

\begin{table}
\begin{tabular}{cc}
\hline
\textbf{Model} & \textbf{$U_{0}$ MAE (meV)}\\ \hline
Cormorant \cite{cormorant} & 22 \\
SEGNN \cite{segnn} & 15 \\
SchNet \cite{schnet} & 14 \\
EGNN \cite{egnn} & 11 \\
Equiformer \cite{equiformer}& 6.6 \\
DimeNet++ \cite{dimenet++} & 6.3 \\
SphereNet \cite{spherenet} & 6.3 \\
ET$_{\mathrm{old}}$ \cite{et} & 6.2 \\
PaiNN \cite{painn} & 5.9 \\
Allegro \cite{allegro} & 4.7 \\
MACE \cite{mace}& 4.1 \\
\hline
ET$_{\mathrm{new}}$ & 5.7 \\
TensorNet 0L & 7.2 \\
TensorNet 1L & 4.7  \\
TensorNet 2L & 4.4 \\
TensorNet 3L* & 3.9 \\
\hline
\end{tabular}
\caption{Mean absolute error in meV for different models trained on QM9 target property $U_0$. TensorNet 3L* uses an embedding dimension of 256, while in other cases 128. For the ET, subscripts new and old correspond to the new and the original implementation, that is, with \texttt{vector\_cutoff = True} and \texttt{False}, respectively.}
\label{tab:qm9r}
\end{table}

We trained the model on the MD17 aspirin dataset (Figure \ref{fig:trainings}) using the hyperparameters defined for the original version of the ET (Table S1
, with the addition of \texttt{vector\_cutoff = True}), giving final test MAEs of $0.139$ kcal/mol and $0.232$ kcal/mol/\AA in energies and forces, respectively, compared to the original implementation which gave $0.123$ kcal/mol and $0.253$ kcal/mol/\AA\cite{et}. Regarding QM9 $U_0$, we reused the original hyperparameters for the dataset found in Table S2
  (again, adding \texttt{vector\_cutoff = True}), and comparative results can be found in Table \ref{tab:qm9r}.


\subsection{Molecular Simulations}

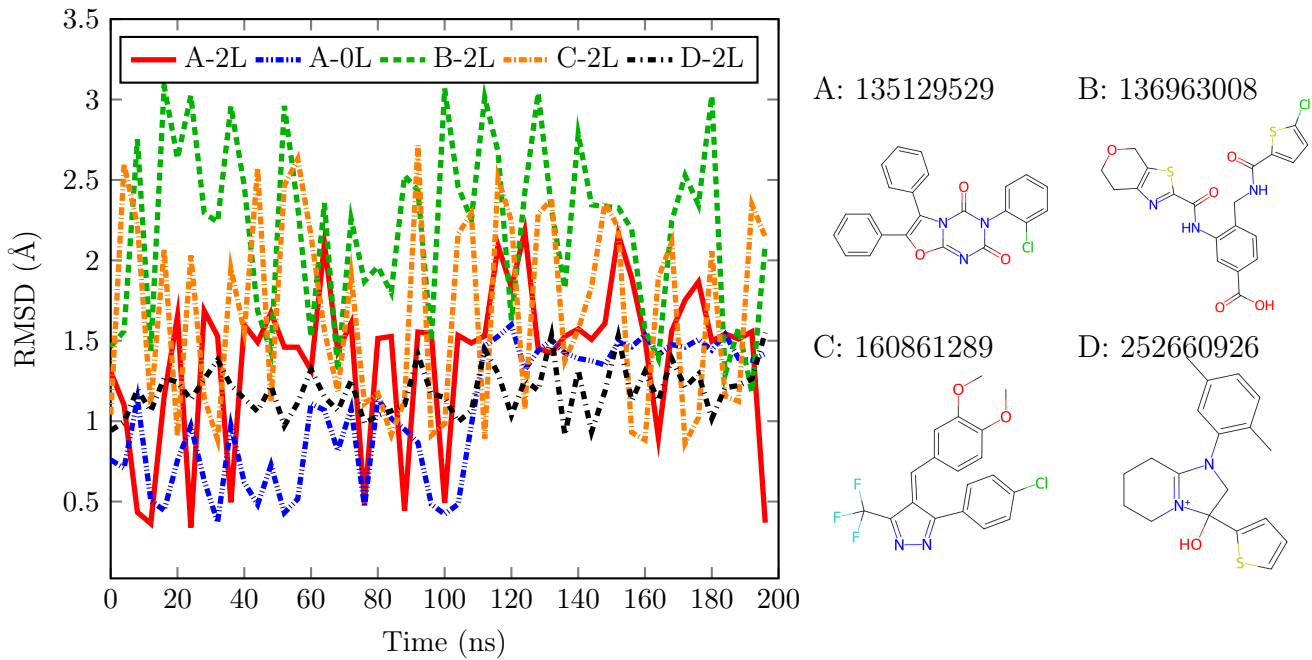
\begin{figure*}
\centering
\begin{minipage}{.59\linewidth}
\centering
\begin{tikzpicture}
  \newcommand{\thickness}{2pt}
\begin{groupplot}[
    title style={at={(0.5,1.2)}},
    xmin=0,
    xmax=200,
    ymax=3.5,
    legend pos=north east,
    legend style={legend columns=-1},
    width=\linewidth,
    cycle list={
      {red, solid, line width=\thickness},
      {blue, densely dashdotdotted, line width=\thickness},
      {black!30!green, densely dashed, line width=\thickness},
      {orange, densely dashdotted, line width=\thickness},
      {black, dashdotted, line width=\thickness}
    },
    each nth point=40,
    axis line style={thick},
    tick style={thick},
    legend style={draw, thick, font=\small},
    legend cell align={left},
    label style={font=\small},
    tick label style={font=\small},
    ]
\nextgroupplot[
    xlabel={Time (ns)},
    ylabel={RMSD (\AA)}
    ]
\addplot table [x=time, y=135129529_2L, col sep=comma] {extended_stability_rmsd.csv};
\addlegendentry{A-2L}
\addplot table [x=time, y=135129529_0L, col sep=comma] {extended_stability_rmsd.csv};
\addlegendentry{A-0L}
\addplot table [x=time, y=136963008_2L, col sep=comma] {extended_stability_rmsd.csv};
\addlegendentry{B-2L}
\addplot table [x=time, y=160861289_2L, col sep=comma] {extended_stability_rmsd.csv};
\addlegendentry{C-2L}
\addplot table [x=time, y=252660926_2L, col sep=comma] {extended_stability_rmsd.csv};
\addlegendentry{D-2L}
\end{groupplot}
\end{tikzpicture}
\end{minipage}\hfill 
\begin{minipage}{.39\linewidth}
\begin{tikzpicture}
  \node (img1) {\includegraphics[width=0.45\textwidth]{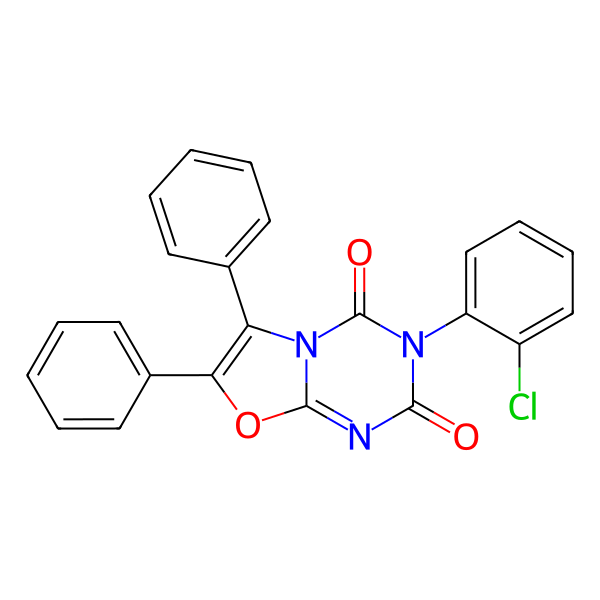}};
  \node[below right, inner sep=0pt, outer sep=0pt] at (img1.north west) {A: 135129529};
\end{tikzpicture}\hfill%
\begin{tikzpicture}
  \node(img2) {\includegraphics[width=0.45\textwidth]{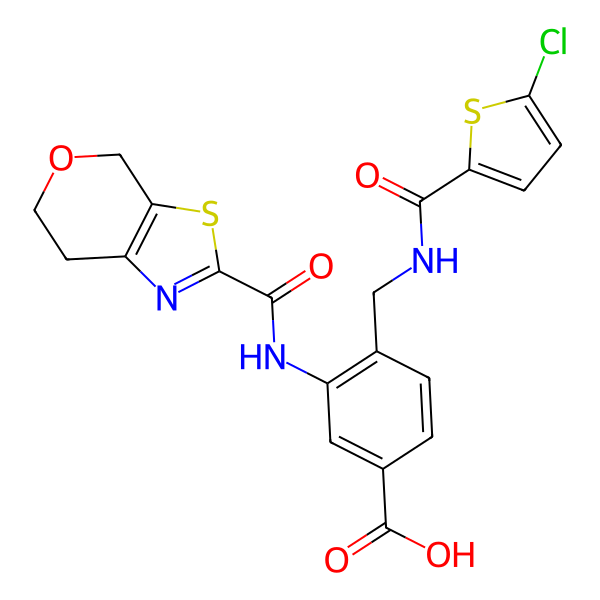}};
  \node[below right, inner sep=0pt, outer sep=0pt] at (img2.north west) {B: 136963008};
\end{tikzpicture}
\begin{tikzpicture}
  \node(img3) {\includegraphics[width=0.45\textwidth]{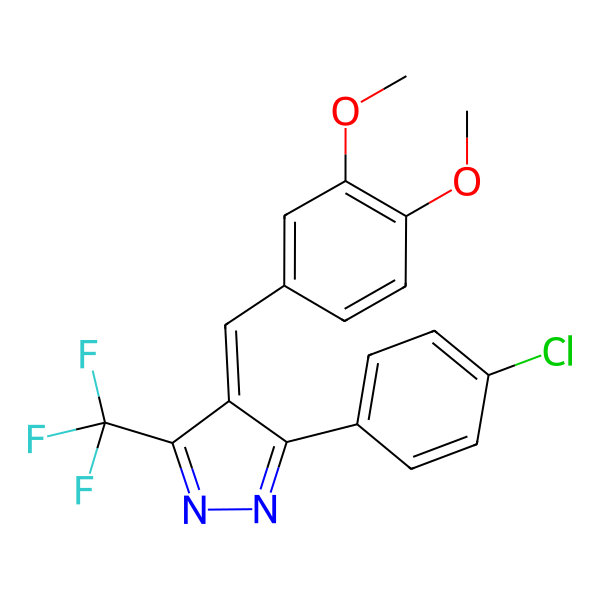}};
  \node[below right, inner sep=0pt, outer sep=0pt] at (img3.north west) {C: 160861289};
\end{tikzpicture}\hfill%
\begin{tikzpicture}
  \node(img4) {\includegraphics[width=0.45\textwidth]{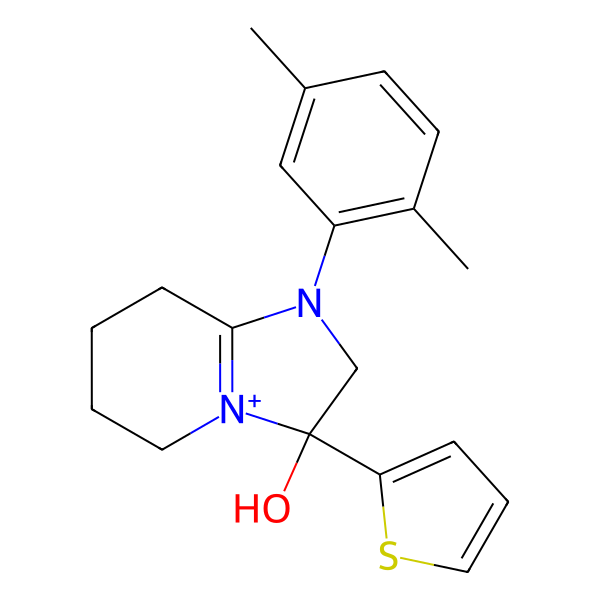}};
  \node[below right, inner sep=0pt, outer sep=0pt] at (img4.north west) {D: 252660926};
\end{tikzpicture}
\end{minipage}
 \caption{(Left) RMSD analysis for the trajectories of 4 molecules outside of the training set. Simulations are carried out with TensorNet 2L, using the parameters in Table S4,
 with the exception of A-0L, in which a 0L TensorNet model is showcased. Presented data is plotted only every 4 ns for visualization clarity. (Right) Representation of the simulated molecules. Labels show the PubChem ID for each molecule.}
 \label{fig:stability-analysis}
\end{figure*}

We performed NVT molecular dynamics simulations, in vacuum, employing TensorNet models trained on the ANI-2x dataset \cite{ani2x}. A table detailing the hyperparameters is provided for reference in Table S4. 
Note that we did not include any physical priors in these trainings nor in the subsequent simulations, i.e. all forces in the system come from the model itself.
Starting from the SPICE dataset \cite{spice}, we selected the PubChem subset and utilized it to create a test set comprising four randomly chosen conformers. Following the order as depicted in Figure \ref{fig:stability-analysis}, the number of atoms for each molecule is 44, 47, 41, and 46. This test set aimed to evaluate the ability of the NNP to perform stable molecular dynamics (MD) simulations on molecules not encountered during the training stage \cite{stable1,stable2}. The training dataset, as well as the PubChem subset, represent a broad diversification of molecules containing the elements H, C, N, O, S, F, Cl. To generate the input data, the SMILES and the coordinates of interest were used to build a molecule object using openff-toolkit \cite{jeff_wagner_2023_10103216}, and the atomic numbers were used as embeddings. Using the more accurate TensorNet 2L model, a $200$ ns trajectory, i.e. $2\cdot 10^8$ steps \footnote{Note a simulation step contains one forward pass of our model to compute total energy in addition to a backward pass to compute atomic forces.}, with a time-step of 1 fs was generated for each molecule using OpenMM's\cite{openmm8} \texttt{LangevinMiddleIntegrator} at $298.5$K and a friction coefficient of 1 ps$^{-1}$. We also used for one of the molecules a TensorNet 0L model with the same simulation settings to test its stability. A root mean square displacement (RMSD) analysis was performed for each trajectory taking the starting conformation as a reference, see Figure \ref{fig:stability-analysis}. The results highlight the model's ability to run stable MD simulations, even for the 0L case where the model's receptive field and parameter count are substantially reduced. 
In terms of computational efficiency, for the specific cases in this section, TensorNet 0L achieves a simulation speed of about 70 ns/day, while the 2-layer version runs at approximately 20 ns/day. For comparison, considering the same number of particles, i.e. around 40 atoms, GFN2-xTB \cite{grimme2017robust} runs at 0.05 ns/day, while PySCF, using B3LYP/6-31g level of theory, \cite{sun2018pyscf} runs at $3 \cdot 10^{-4}$ ns/day.
\subsection{Speed performance}
All results presented in this work were carried out using an NVIDIA RTX4090 GPU (driver version 525.147) with a dual 8 core Intel(R) Xeon(R) Silver 4110 CPU in Ubuntu 20.04. We used CUDA 12.0 with Pytorch version 2.1.0 from conda-forge. We provide all timings in million steps per day, which can be easily converted to nanoseconds per day. These units are more commonly used in molecular dynamics settings, and the conversion can be done by taking the quantity in million steps per day times the timestep in femtoseconds. Therefore, for example, 1 million steps per day is equivalent to 1 ns/day for a timestep of 1 fs.

\begin{figure}
\begin{tikzpicture}
  \begin{groupplot}[
    group style={
      group size=1 by 3,
      xlabels at=edge bottom,
      ylabels at=edge left,
      vertical sep=0,
      horizontal sep=0
    },
    xmode=log, ymode=log,
    grid=major,
    legend pos=south west,
    width=\linewidth,
    cycle list={
      {black, line width=1.5pt},
      {red, line width=1.5pt},
      {blue, line width=1.5pt},
      {green, line width=1.5pt},
    },
    xlabel={Number of particles},
    ylabel={Million steps per day},
    tick label style={font=\small}
  ]
  \nextgroupplot[xticklabel=\empty]
  \node[anchor=north west, fill=white, font=\large] at (axis description cs:0.75,0.9) {0L};

  \addplot table [x=nparticles, y expr=86.4/\thisrow{0L}, col sep=comma] {tn_inference_benchmark/tn_plain.csv};
  \addplot table [x=nparticles, y expr=86.4/\thisrow{0L}, col sep=comma] {tn_inference_benchmark/tn_compile2.csv};
  \addplot table [x=nparticles, y expr=86.4/\thisrow{0L}, col sep=comma] {tn_inference_benchmark/tn_graphs.csv};
  
  \nextgroupplot[xticklabel=\empty]
 \node[anchor=north west, fill=white, font=\large] at (axis description cs:0.75,0.9) {1L};

  \addplot table [x=nparticles, y expr=86.4/\thisrow{1L}, col sep=comma] {tn_inference_benchmark/tn_plain.csv};
  \addplot table [x=nparticles, y expr=86.4/\thisrow{1L}, col sep=comma] {tn_inference_benchmark/tn_compile2.csv};
  \addplot table [x=nparticles, y expr=86.4/\thisrow{1L}, col sep=comma] {tn_inference_benchmark/tn_graphs.csv};

  \nextgroupplot[ymax=200]
    \node[anchor=north west, fill=white, font=\large] at (axis description cs:0.75,0.9) {2L};

  \addplot table [x=nparticles, y expr=86.4/\thisrow{2L}, col sep=comma] {tn_inference_benchmark/tn_plain.csv};
\addlegendentry{Plain}
  \addplot table [x=nparticles, y expr=86.4/\thisrow{2L}, col sep=comma] {tn_inference_benchmark/tn_compile2.csv};
    \addlegendentry{Compile}
  \addplot table [x=nparticles, y expr=86.4/\thisrow{2L}, col sep=comma] {tn_inference_benchmark/tn_graphs.csv};
    \addlegendentry{Graph}

  \end{groupplot}
\end{tikzpicture}
\caption{Comparison between TensorNet inference times (energy and forces) with $0$, $1$ and $2$ interaction layers, embedding dimension $128$, $64$ neighbors on average. All atoms are passed in a single batch. Plain represents the bare TensorNet implementation; with Compile the module has been preprocessed with \texttt{torch.compile} with the "max-autotune" option; for Graphs the whole computation has been captured into a single CUDA graph.}
\label{fig:inference_times}
\end{figure}
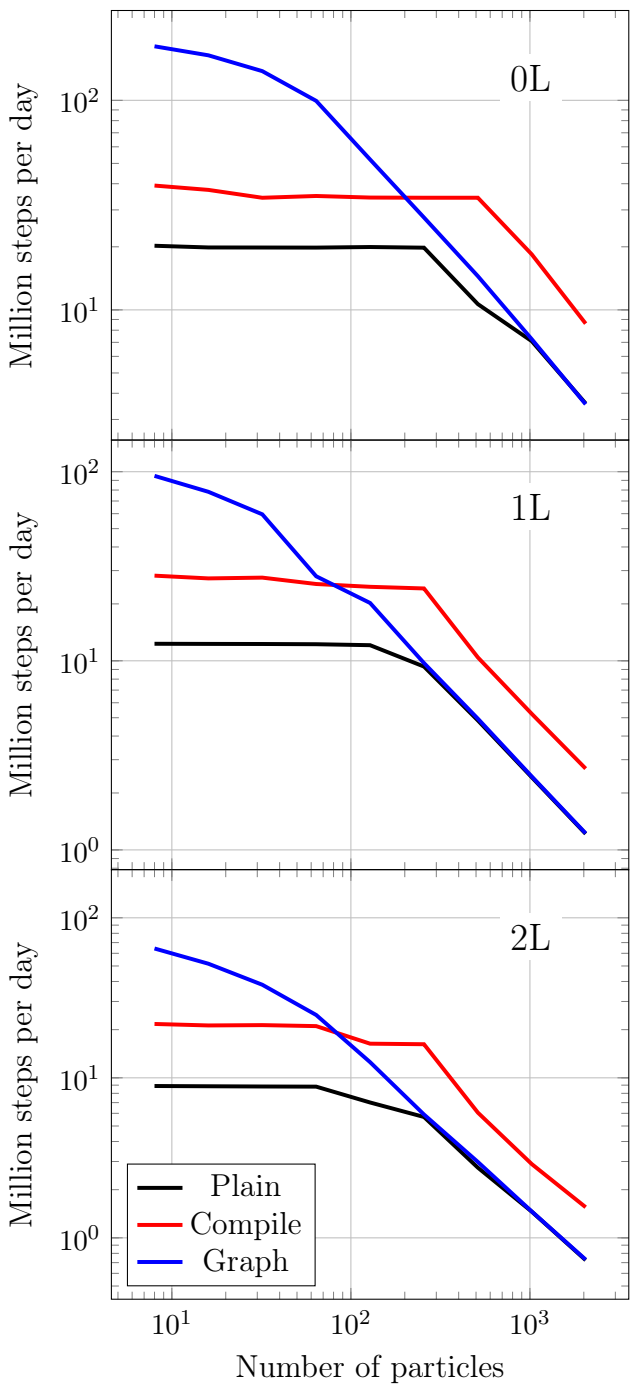

To study the optimization strategies laid out in section \ref{sec:opt} we show energy and forces inference performance for several equivalent implementations of TensorNet in Figure \ref{fig:inference_times}. Note that in TorchMD-Net, running inference requires one backpropagation step to compute forces as the negative gradient of the energies with respect to the input positions, which are computed via Autograd. This step is also included in these benchmarks.
We also explore inference times for some molecules with varying numbers of atoms in Table \ref{tab:prot}. For these molecules, which can be found in the repository for speed benchmarking purposes, we measure time to compute the potential energy and atomic forces of a single example using TensorNet with 0, 1 and 2 interaction layers. Again, we express this time in million steps per day. In all cases, we use a cutoff of $4.5$\AA, an embedding dimension of $128$, $32$ radial basis functions and a maximum of $32$ neighbors per particle. 
\begin{table*}
\centering
\begin{tabular}{cccccccccc}
\hline
\textbf{Molecule (atoms)} & \textbf{P 0L} & \textbf{P 1L} & \textbf{P 2L} & \textbf{C 0L} & \textbf{C 1L} & \textbf{C 2L} & \textbf{G 0L} & \textbf{G 1L} & \textbf{G 2L} \\ 
\hline
Alanine dipeptide (22) & 19.86 & 10.29 & 8.50 & 40.19 & 28.70 & 21.23 & 172.80 & 84.71 & 56.47 \\
Testosterone (49) & 15.05 & 11.93 & 8.56 & 38.57 & 27.00 & 21.49 & 154.29 & 63.53 & 39.82\\
Chignolin (166) & 19.77 & 11.88 & 7.90 & 36.77 & 24.90 & 21.39 & 77.14 & 26.02 & 15.57\\
DHFR (2489) & 5.56 & 1.67 & 0.98 & 14.47 & 3.27 & 1.83 & 5.65 & 1.69 & 1.00\\
Factor IX (5807) & 2.32 & 0.69 & 0.41 & 5.42 & 1.35 & 0.77 & 2.33 & 0.70 & 0.42\\
\hline
\end{tabular}
\caption{TensorNet inference times in million steps per day for the "Plain" (\textbf{P}), "Compile" (\textbf{C}) and "Graph" (\textbf{G}) implementations and varying number of message passing layers L}.
\label{tab:prot}
\end{table*}
We make sure not to include any warmup times in these benchmarks by running the models for $100$ iterations before timing. We refer as "Graph" to an implementation that has been modified to ensure every CUDA graph requirement is met. For "Compile" the implementation is carefully tailored to look for the best performance in \texttt{torch.compile} in addition to the changes introduced for "Graph". Finally, "Plain" represents the baseline implementation in PyTorch.

Although in principle the code received by the compiler is entirely capturable by a graph, it often decides to capture only some sections of it, introducing other kinds of optimizations instead. This is also made evident by the appearance of the same kind of "plateau" performance for smaller workloads in both Plain and Compile, which can be attributed to a bottleneck produced by kernel launch overhead. Still, the torch compiler is able to provide a speedup of a factor of 2 to 3 for all workloads with respect to the original implementation.

CUDA kernel overhead (and thus the performance gain of CUDA graphs) is expected to dominate for small workloads, where it is usual for the kernel launching time to be larger than the actual execution. Figure \ref{fig:inference_times} indeed corroborates this by showing speedups between 10 and 2 times for molecules with up to a few hundred atoms and for all numbers of interaction layers (0, 1, and 2). Starting from workloads consisting of several hundreds of atoms, the performance of the Plain version is recovered.

Another factor to take into account is the additional memory consumption introduced by enforcing static shapes. There are two sources of memory overhead in this case; the padding of the edges until a maximum number of pairs and the extra node (ghost atom) that these edges are assigned to. The extra node bears a cost scaling with $1/N$ and is thus negligible. On the other hand, the extra edges have a cost in memory that scales with the difference between the maximum number of pairs and the actual pairs found. In practice these values are chosen to be as close as possible while leaving a safety buffer, making the expected memory overhead a small percentage of the total memory except in some pathological cases\footnote{For instance, for simplicity we fix the maximum number of pairs as 32 times the number of atoms for the benchmarks in table \ref{tab:prot} even for Alanine dipeptide, which only has 22 atoms.}.

\section{Conclusions}

TorchMD-Net has significantly evolved in its recent iterations, becoming a comprehensive platform for neural network potentials (NNPs). It provides researchers with robust tools for both rapid prototyping of new models and executing production-level tasks. However, despite these advancements, NNPs still face substantial challenges before they can fully replace traditional force fields in molecular dynamics simulations. Currently, while the necessary software infrastructure is largely in place, as evidenced by the first-class support for NNPs in popular packages \cite{openmm8}, issues such as memory requirements and computational performance remain significant concerns.

The impact of memory limitations is anticipated to diminish with ongoing hardware advancements. Yet, enhancing computational performance to a level that is competitive with traditional methods necessitates more intricate strategies. This involves developing architectures and their implementations in a manner that leverages the full capabilities of GPU hardware. 

From a software development perspective, the compilation functionality within PyTorch is an evolving feature, still in its early stages. Its current development trajectory, which aims to minimize the necessary code modifications for effective utilization, suggests that future PyTorch releases will likely bring performance enhancements. Continuous improvements in the relevant toolset, encompassing PyTorch, CUDA, Triton, and others, are gradually narrowing the performance gap between highly optimized code and more straightforward implementations.

\section*{Acknowledgement}
We thank Prof. Jan Rezac for discovering the spurious discontinuity in the Equivariant Transformer.
G. S. is financially supported by Generalitat de Catalunya's Agency for Management of University and Research Grants (AGAUR) PhD grant FI-2-00587. This project has received funding from
the European Union’s Horizon 2020 research and innovation programme under grant agreement No. 823712;
and the project PID2020-116564GB-I00 has been funded by MCIN / AEI / 10.13039/501100011033; 
Research reported in this publication was supported by the National Institute of General Medical Sciences (NIGMS) of the National Institutes of Health under award number R01GM140090. 
The content is solely the responsibility of the authors and does not necessarily represent the official views of the National Institutes of Health.
\section*{ASSOCIATED CONTENT}
Supporting Information available. Hyperparameters used for the different trainings presented throughout the manuscript.
\bibliography{references}

\providecommand{\latin}[1]{#1}
\makeatletter
\providecommand{\doi}
  {\begingroup\let\do\@makeother\dospecials
  \catcode`\{=1 \catcode`\}=2 \doi@aux}
\providecommand{\doi@aux}[1]{\endgroup\texttt{#1}}
\makeatother
\providecommand*\mcitethebibliography{\thebibliography}
\csname @ifundefined\endcsname{endmcitethebibliography}  {\let\endmcitethebibliography\endthebibliography}{}
\begin{mcitethebibliography}{72}
\providecommand*\natexlab[1]{#1}
\providecommand*\mciteSetBstSublistMode[1]{}
\providecommand*\mciteSetBstMaxWidthForm[2]{}
\providecommand*\mciteBstWouldAddEndPuncttrue
  {\def\EndOfBibitem{\unskip.}}
\providecommand*\mciteBstWouldAddEndPunctfalse
  {\let\EndOfBibitem\relax}
\providecommand*\mciteSetBstMidEndSepPunct[3]{}
\providecommand*\mciteSetBstSublistLabelBeginEnd[3]{}
\providecommand*\EndOfBibitem{}
\mciteSetBstSublistMode{f}
\mciteSetBstMaxWidthForm{subitem}{(\alph{mcitesubitemcount})}
\mciteSetBstSublistLabelBeginEnd
  {\mcitemaxwidthsubitemform\space}
  {\relax}
  {\relax}

\bibitem[Behler and Parrinello(2007)Behler, and Parrinello]{behlerparrinello}
Behler,~J.; Parrinello,~M. Generalized Neural-Network Representation of High-Dimensional Potential-Energy Surfaces. \emph{Phys. Rev. Lett.} \textbf{2007}, \emph{98}, 146401\relax
\mciteBstWouldAddEndPuncttrue
\mciteSetBstMidEndSepPunct{\mcitedefaultmidpunct}
{\mcitedefaultendpunct}{\mcitedefaultseppunct}\relax
\EndOfBibitem
\bibitem[Kocer \latin{et~al.}(2021)Kocer, Ko, and Behler]{nnp1}
Kocer,~E.; Ko,~T.~W.; Behler,~J. Neural Network Potentials: A Concise Overview of Methods. arXiv, 2021; 10.48550/arXiv.2107.03727\relax
\mciteBstWouldAddEndPuncttrue
\mciteSetBstMidEndSepPunct{\mcitedefaultmidpunct}
{\mcitedefaultendpunct}{\mcitedefaultseppunct}\relax
\EndOfBibitem
\bibitem[Behler(2016)]{nnp2}
Behler,~J. {Perspective: Machine learning potentials for atomistic simulations}. \emph{The Journal of Chemical Physics} \textbf{2016}, \emph{145}, 170901\relax
\mciteBstWouldAddEndPuncttrue
\mciteSetBstMidEndSepPunct{\mcitedefaultmidpunct}
{\mcitedefaultendpunct}{\mcitedefaultseppunct}\relax
\EndOfBibitem
\bibitem[Schütt \latin{et~al.}(2018)Schütt, Sauceda, Kindermans, Tkatchenko, and Müller]{nnp3}
Schütt,~K.~T.; Sauceda,~H.~E.; Kindermans,~P.-J.; Tkatchenko,~A.; Müller,~K.-R. {SchNet – A deep learning architecture for molecules and materials}. \emph{The Journal of Chemical Physics} \textbf{2018}, \emph{148}, 241722\relax
\mciteBstWouldAddEndPuncttrue
\mciteSetBstMidEndSepPunct{\mcitedefaultmidpunct}
{\mcitedefaultendpunct}{\mcitedefaultseppunct}\relax
\EndOfBibitem
\bibitem[Deringer \latin{et~al.}(2019)Deringer, Caro, and Csányi]{nnp4}
Deringer,~V.~L.; Caro,~M.~A.; Csányi,~G. Machine Learning Interatomic Potentials as Emerging Tools for Materials Science. \emph{Advanced Materials} \textbf{2019}, \emph{31}, 1902765\relax
\mciteBstWouldAddEndPuncttrue
\mciteSetBstMidEndSepPunct{\mcitedefaultmidpunct}
{\mcitedefaultendpunct}{\mcitedefaultseppunct}\relax
\EndOfBibitem
\bibitem[Botu \latin{et~al.}(2017)Botu, Batra, Chapman, and Ramprasad]{nnp5}
Botu,~V.; Batra,~R.; Chapman,~J.; Ramprasad,~R. Machine Learning Force Fields: Construction, Validation, and Outlook. \emph{The Journal of Physical Chemistry C} \textbf{2017}, \emph{121}, 511--522\relax
\mciteBstWouldAddEndPuncttrue
\mciteSetBstMidEndSepPunct{\mcitedefaultmidpunct}
{\mcitedefaultendpunct}{\mcitedefaultseppunct}\relax
\EndOfBibitem
\bibitem[Ko \latin{et~al.}(2021)Ko, Finkler, Goedecker, and Behler]{nnp6}
Ko,~T.~W.; Finkler,~J.~A.; Goedecker,~S.; Behler,~J. A fourth-generation high-dimensional neural network potential with accurate electrostatics including non-local charge transfer. \emph{Nature Communications} \textbf{2021}, \emph{12}, 398\relax
\mciteBstWouldAddEndPuncttrue
\mciteSetBstMidEndSepPunct{\mcitedefaultmidpunct}
{\mcitedefaultendpunct}{\mcitedefaultseppunct}\relax
\EndOfBibitem
\bibitem[Sch\"{u}tt \latin{et~al.}(2023)Sch\"{u}tt, Hessmann, Gebauer, Lederer, and Gastegger]{schnetpack}
Sch\"{u}tt,~K.~T.; Hessmann,~S. S.~P.; Gebauer,~N. W.~A.; Lederer,~J.; Gastegger,~M. SchNetPack 2.0: A neural network toolbox for atomistic machine learning. \emph{The Journal of Chemical Physics} \textbf{2023}, \emph{158}\relax
\mciteBstWouldAddEndPuncttrue
\mciteSetBstMidEndSepPunct{\mcitedefaultmidpunct}
{\mcitedefaultendpunct}{\mcitedefaultseppunct}\relax
\EndOfBibitem
\bibitem[Gao \latin{et~al.}(2020)Gao, Ramezanghorbani, Isayev, Smith, and Roitberg]{torchani}
Gao,~X.; Ramezanghorbani,~F.; Isayev,~O.; Smith,~J.~S.; Roitberg,~A.~E. TorchANI: A Free and Open Source PyTorch-Based Deep Learning Implementation of the ANI Neural Network Potentials. \emph{Journal of Chemical Information and Modeling} \textbf{2020}, \emph{60}, 3408–3415\relax
\mciteBstWouldAddEndPuncttrue
\mciteSetBstMidEndSepPunct{\mcitedefaultmidpunct}
{\mcitedefaultendpunct}{\mcitedefaultseppunct}\relax
\EndOfBibitem
\bibitem[Zeng \latin{et~al.}(2023)Zeng, Zhang, Lu, Mo, Li, Chen, Rynik, Huang, Li, Shi, Wang, Ye, Tuo, Yang, Ding, Li, Tisi, Zeng, Bao, Xia, Huang, Muraoka, Wang, Chang, Yuan, Bore, Cai, Lin, Wang, Xu, Zhu, Luo, Zhang, Goodall, Liang, Singh, Yao, Zhang, Wentzcovitch, Han, Liu, Jia, York, E, Car, Zhang, and Wang]{deepmd}
Zeng,~J.; Zhang,~D.; Lu,~D.; Mo,~P.; Li,~Z.; Chen,~Y.; Rynik,~M.; Huang,~L.; Li,~Z.; Shi,~S. \latin{et~al.}  DeePMD-kit v2: A software package for deep potential models. \emph{The Journal of Chemical Physics} \textbf{2023}, \emph{159}\relax
\mciteBstWouldAddEndPuncttrue
\mciteSetBstMidEndSepPunct{\mcitedefaultmidpunct}
{\mcitedefaultendpunct}{\mcitedefaultseppunct}\relax
\EndOfBibitem
\bibitem[Th\"{o}lke and De~Fabritiis(2022)Th\"{o}lke, and De~Fabritiis]{et}
Th\"{o}lke,~P.; De~Fabritiis,~G. TorchMD-NET: Equivariant Transformers for Neural Network based Molecular Potentials. arXiv, 2022; 10.48550/ARXIV.2202.02541\relax
\mciteBstWouldAddEndPuncttrue
\mciteSetBstMidEndSepPunct{\mcitedefaultmidpunct}
{\mcitedefaultendpunct}{\mcitedefaultseppunct}\relax
\EndOfBibitem
\bibitem[Majewski \latin{et~al.}(2022)Majewski, Pérez, Thölke, Doerr, Charron, Giorgino, Husic, Clementi, Noé, and Fabritiis]{majewski2022machine}
Majewski,~M.; Pérez,~A.; Thölke,~P.; Doerr,~S.; Charron,~N.~E.; Giorgino,~T.; Husic,~B.~E.; Clementi,~C.; Noé,~F.; Fabritiis,~G.~D. Machine Learning Coarse-Grained Potentials of Protein Thermodynamics. arXiv, 2022; 10.48550/arXiv.2212.07492\relax
\mciteBstWouldAddEndPuncttrue
\mciteSetBstMidEndSepPunct{\mcitedefaultmidpunct}
{\mcitedefaultendpunct}{\mcitedefaultseppunct}\relax
\EndOfBibitem
\bibitem[Simeon and De~Fabritiis(2023)Simeon, and De~Fabritiis]{simeon2023tensornet}
Simeon,~G.; De~Fabritiis,~G. TensorNet: Cartesian Tensor Representations for Efficient Learning of Molecular Potentials. Advances in Neural Information Processing Systems. 2023; pp 37334--37353\relax
\mciteBstWouldAddEndPuncttrue
\mciteSetBstMidEndSepPunct{\mcitedefaultmidpunct}
{\mcitedefaultendpunct}{\mcitedefaultseppunct}\relax
\EndOfBibitem
\bibitem[Paszke \latin{et~al.}(2019)Paszke, Gross, Massa, Lerer, Bradbury, Chanan, Killeen, Lin, Gimelshein, Antiga, Desmaison, Kopf, Yang, DeVito, Raison, Tejani, Chilamkurthy, Steiner, Fang, Bai, and Chintala]{pytorch}
Paszke,~A.; Gross,~S.; Massa,~F.; Lerer,~A.; Bradbury,~J.; Chanan,~G.; Killeen,~T.; Lin,~Z.; Gimelshein,~N.; Antiga,~L. \latin{et~al.}  \emph{Advances in Neural Information Processing Systems 32}; Curran Associates, Inc., 2019; pp 8024--8035\relax
\mciteBstWouldAddEndPuncttrue
\mciteSetBstMidEndSepPunct{\mcitedefaultmidpunct}
{\mcitedefaultendpunct}{\mcitedefaultseppunct}\relax
\EndOfBibitem
\bibitem[lig()]{lightning}
Pytorch Lightning. \url{lightning.ai/pytorch-lightning}, (Accessed February 15, 2024)\relax
\mciteBstWouldAddEndPuncttrue
\mciteSetBstMidEndSepPunct{\mcitedefaultmidpunct}
{\mcitedefaultendpunct}{\mcitedefaultseppunct}\relax
\EndOfBibitem
\bibitem[conda-forge community(2015)]{conda_forge}
conda-forge community {The conda-forge Project: Community-based Software Distribution Built on the conda Package Format and Ecosystem}. 2015; \url{https://doi.org/10.5281/zenodo.4774216}\relax
\mciteBstWouldAddEndPuncttrue
\mciteSetBstMidEndSepPunct{\mcitedefaultmidpunct}
{\mcitedefaultendpunct}{\mcitedefaultseppunct}\relax
\EndOfBibitem
\bibitem[tor()]{torchmdnetdocs}
TorchMD-NET Documentation. torchmd-net.readthedocs.io, (Accessed February 13, 2024)\relax
\mciteBstWouldAddEndPuncttrue
\mciteSetBstMidEndSepPunct{\mcitedefaultmidpunct}
{\mcitedefaultendpunct}{\mcitedefaultseppunct}\relax
\EndOfBibitem
\bibitem[Biersack and Ziegler(1982)Biersack, and Ziegler]{zblprior}
Biersack,~J.~P.; Ziegler,~J.~F. \emph{Ion Implantation Techniques}; Springer Berlin Heidelberg, 1982; p 122–156\relax
\mciteBstWouldAddEndPuncttrue
\mciteSetBstMidEndSepPunct{\mcitedefaultmidpunct}
{\mcitedefaultendpunct}{\mcitedefaultseppunct}\relax
\EndOfBibitem
\bibitem[Eastman \latin{et~al.}(2024)Eastman, Galvelis, Peláez, Abreu, Farr, Gallicchio, Gorenko, Henry, Hu, Huang, Krämer, Michel, Mitchell, Pande, Rodrigues, Rodriguez-Guerra, Simmonett, Singh, Swails, Turner, Wang, Zhang, Chodera, De~Fabritiis, and Markland]{openmm8}
Eastman,~P.; Galvelis,~R.; Peláez,~R.~P.; Abreu,~C. R.~A.; Farr,~S.~E.; Gallicchio,~E.; Gorenko,~A.; Henry,~M.~M.; Hu,~F.; Huang,~J. \latin{et~al.}  OpenMM 8: Molecular Dynamics Simulation with Machine Learning Potentials. \emph{The Journal of Physical Chemistry B} \textbf{2024}, \emph{128}, 109--116, PMID: 38154096\relax
\mciteBstWouldAddEndPuncttrue
\mciteSetBstMidEndSepPunct{\mcitedefaultmidpunct}
{\mcitedefaultendpunct}{\mcitedefaultseppunct}\relax
\EndOfBibitem
\bibitem[ope()]{openmmtorch}
OpenMM-Torch. \url{https://github.com/openmm/openmm-torch }, (Accessed: February 15 2024)\relax
\mciteBstWouldAddEndPuncttrue
\mciteSetBstMidEndSepPunct{\mcitedefaultmidpunct}
{\mcitedefaultendpunct}{\mcitedefaultseppunct}\relax
\EndOfBibitem
\bibitem[Simeon \latin{et~al.}(2024)Simeon, Mirarchi, Pelaez, Galvelis, and De~Fabritiis]{qs}
Simeon,~G.; Mirarchi,~A.; Pelaez,~R.~P.; Galvelis,~R.; De~Fabritiis,~G. On the Inclusion of Charge and Spin States in Cartesian Tensor Neural Network Potentials. arXiv, 2024; 10.48550/ARXIV.2403.15073\relax
\mciteBstWouldAddEndPuncttrue
\mciteSetBstMidEndSepPunct{\mcitedefaultmidpunct}
{\mcitedefaultendpunct}{\mcitedefaultseppunct}\relax
\EndOfBibitem
\bibitem[Bronstein \latin{et~al.}(2021)Bronstein, Bruna, Cohen, and Veličković]{messagepassing1}
Bronstein,~M.~M.; Bruna,~J.; Cohen,~T.; Veličković,~P. Geometric Deep Learning: Grids, Groups, Graphs, Geodesics, and Gauges. arXiv, 2021; 10.48550/ARXIV.2104.13478\relax
\mciteBstWouldAddEndPuncttrue
\mciteSetBstMidEndSepPunct{\mcitedefaultmidpunct}
{\mcitedefaultendpunct}{\mcitedefaultseppunct}\relax
\EndOfBibitem
\bibitem[Gilmer \latin{et~al.}(2017)Gilmer, Schoenholz, Riley, Vinyals, and Dahl]{messagepassing2}
Gilmer,~J.; Schoenholz,~S.~S.; Riley,~P.~F.; Vinyals,~O.; Dahl,~G.~E. Neural Message Passing for Quantum Chemistry. arXiv, 2017; 10.48550/ARXIV.1704.01212\relax
\mciteBstWouldAddEndPuncttrue
\mciteSetBstMidEndSepPunct{\mcitedefaultmidpunct}
{\mcitedefaultendpunct}{\mcitedefaultseppunct}\relax
\EndOfBibitem
\bibitem[Joshi \latin{et~al.}(2023)Joshi, Bodnar, Mathis, Cohen, and Liò]{expressive}
Joshi,~C.~K.; Bodnar,~C.; Mathis,~S.~V.; Cohen,~T.; Liò,~P. On the Expressive Power of Geometric Graph Neural Networks. arXiv, 2023; \url{https://arxiv.org/abs/2301.09308}, 10.48550/ARXIV.2301.09308\relax
\mciteBstWouldAddEndPuncttrue
\mciteSetBstMidEndSepPunct{\mcitedefaultmidpunct}
{\mcitedefaultendpunct}{\mcitedefaultseppunct}\relax
\EndOfBibitem
\bibitem[Duval \latin{et~al.}(2023)Duval, Mathis, Joshi, Schmidt, Miret, Malliaros, Cohen, Lio, Bengio, and Bronstein]{hitchhiker}
Duval,~A.; Mathis,~S.~V.; Joshi,~C.~K.; Schmidt,~V.; Miret,~S.; Malliaros,~F.~D.; Cohen,~T.; Lio,~P.; Bengio,~Y.; Bronstein,~M. A Hitchhiker's Guide to Geometric GNNs for 3D Atomic Systems. arXiv, 2023; 10.48550/ARXIV.2312.07511\relax
\mciteBstWouldAddEndPuncttrue
\mciteSetBstMidEndSepPunct{\mcitedefaultmidpunct}
{\mcitedefaultendpunct}{\mcitedefaultseppunct}\relax
\EndOfBibitem
\bibitem[Geiger and Smidt(2022)Geiger, and Smidt]{e3nn}
Geiger,~M.; Smidt,~T. e3nn: Euclidean Neural Networks. arXiv, 2022; 10.48550/ARXIV.2207.09453\relax
\mciteBstWouldAddEndPuncttrue
\mciteSetBstMidEndSepPunct{\mcitedefaultmidpunct}
{\mcitedefaultendpunct}{\mcitedefaultseppunct}\relax
\EndOfBibitem
\bibitem[Batzner \latin{et~al.}(2022)Batzner, Musaelian, Sun, Geiger, Mailoa, Kornbluth, Molinari, Smidt, and Kozinsky]{nequip}
Batzner,~S.; Musaelian,~A.; Sun,~L.; Geiger,~M.; Mailoa,~J.~P.; Kornbluth,~M.; Molinari,~N.; Smidt,~T.~E.; Kozinsky,~B. E(3)-equivariant graph neural networks for data-efficient and accurate interatomic potentials. \emph{Nature Communications} \textbf{2022}, \emph{13}\relax
\mciteBstWouldAddEndPuncttrue
\mciteSetBstMidEndSepPunct{\mcitedefaultmidpunct}
{\mcitedefaultendpunct}{\mcitedefaultseppunct}\relax
\EndOfBibitem
\bibitem[Musaelian \latin{et~al.}(2023)Musaelian, Batzner, Johansson, Sun, Owen, Kornbluth, and Kozinsky]{allegro}
Musaelian,~A.; Batzner,~S.; Johansson,~A.; Sun,~L.; Owen,~C.~J.; Kornbluth,~M.; Kozinsky,~B. Learning local equivariant representations for large-scale atomistic dynamics. \emph{Nature Communications} \textbf{2023}, \emph{14}\relax
\mciteBstWouldAddEndPuncttrue
\mciteSetBstMidEndSepPunct{\mcitedefaultmidpunct}
{\mcitedefaultendpunct}{\mcitedefaultseppunct}\relax
\EndOfBibitem
\bibitem[Batatia \latin{et~al.}(2022)Batatia, Kovacs, Simm, Ortner, and Csanyi]{mace}
Batatia,~I.; Kovacs,~D.~P.; Simm,~G. N.~C.; Ortner,~C.; Csanyi,~G. {MACE}: Higher Order Equivariant Message Passing Neural Networks for Fast and Accurate Force Fields. Advances in Neural Information Processing Systems. 2022\relax
\mciteBstWouldAddEndPuncttrue
\mciteSetBstMidEndSepPunct{\mcitedefaultmidpunct}
{\mcitedefaultendpunct}{\mcitedefaultseppunct}\relax
\EndOfBibitem
\bibitem[Sch\"{u}tt \latin{et~al.}(2021)Sch\"{u}tt, Unke, and Gastegger]{painn}
Sch\"{u}tt,~K.~T.; Unke,~O.~T.; Gastegger,~M. Equivariant message passing for the prediction of tensorial properties and molecular spectra. arXiv, 2021; \url{https://arxiv.org/abs/2102.03150}, 10.48550/ARXIV.2102.03150\relax
\mciteBstWouldAddEndPuncttrue
\mciteSetBstMidEndSepPunct{\mcitedefaultmidpunct}
{\mcitedefaultendpunct}{\mcitedefaultseppunct}\relax
\EndOfBibitem
\bibitem[Satorras \latin{et~al.}(2021)Satorras, Hoogeboom, and Welling]{egnn}
Satorras,~V.~G.; Hoogeboom,~E.; Welling,~M. E(n) Equivariant Graph Neural Networks. arXiv, 2021; 10.48550/ARXIV.2102.09844\relax
\mciteBstWouldAddEndPuncttrue
\mciteSetBstMidEndSepPunct{\mcitedefaultmidpunct}
{\mcitedefaultendpunct}{\mcitedefaultseppunct}\relax
\EndOfBibitem
\bibitem[Bihani \latin{et~al.}(2023)Bihani, Pratiush, Mannan, Du, Chen, Miret, Micoulaut, Smedskjaer, Ranu, and Krishnan]{egraffbench}
Bihani,~V.; Pratiush,~U.; Mannan,~S.; Du,~T.; Chen,~Z.; Miret,~S.; Micoulaut,~M.; Smedskjaer,~M.~M.; Ranu,~S.; Krishnan,~N. M.~A. EGraFFBench: Evaluation of Equivariant Graph Neural Network Force Fields for Atomistic Simulations. arXiv, 2023; 10.48550/ARXIV.2310.02428\relax
\mciteBstWouldAddEndPuncttrue
\mciteSetBstMidEndSepPunct{\mcitedefaultmidpunct}
{\mcitedefaultendpunct}{\mcitedefaultseppunct}\relax
\EndOfBibitem
\bibitem[Sch{\"u}tt \latin{et~al.}(2017)Sch{\"u}tt, Arbabzadah, Chmiela, M{\"u}ller, and Tkatchenko]{Schutt2017}
Sch{\"u}tt,~K.~T.; Arbabzadah,~F.; Chmiela,~S.; M{\"u}ller,~K.~R.; Tkatchenko,~A. Quantum-chemical insights from deep tensor neural networks. \emph{Nature Communications} \textbf{2017}, \emph{8}, 13890\relax
\mciteBstWouldAddEndPuncttrue
\mciteSetBstMidEndSepPunct{\mcitedefaultmidpunct}
{\mcitedefaultendpunct}{\mcitedefaultseppunct}\relax
\EndOfBibitem
\bibitem[Unke and Meuwly(2019)Unke, and Meuwly]{Unke2019-tb}
Unke,~O.~T.; Meuwly,~M. {PhysNet}: A Neural Network for Predicting Energies, Forces, Dipole Moments, and Partial Charges. \emph{J Chem Theory Comput} \textbf{2019}, \emph{15}, 3678--3693\relax
\mciteBstWouldAddEndPuncttrue
\mciteSetBstMidEndSepPunct{\mcitedefaultmidpunct}
{\mcitedefaultendpunct}{\mcitedefaultseppunct}\relax
\EndOfBibitem
\bibitem[Unke and Meuwly(2019)Unke, and Meuwly]{physnet}
Unke,~O.~T.; Meuwly,~M. {PhysNet}: A Neural Network for Predicting Energies, Forces, Dipole Moments, and Partial Charges. \emph{Journal of Chemical Theory and Computation} \textbf{2019}, \emph{15}, 3678--3693\relax
\mciteBstWouldAddEndPuncttrue
\mciteSetBstMidEndSepPunct{\mcitedefaultmidpunct}
{\mcitedefaultendpunct}{\mcitedefaultseppunct}\relax
\EndOfBibitem
\bibitem[Unke \latin{et~al.}(2021)Unke, Chmiela, Gastegger, Sch\"{u}tt, Sauceda, and M\"{u}ller]{spookynet}
Unke,~O.~T.; Chmiela,~S.; Gastegger,~M.; Sch\"{u}tt,~K.~T.; Sauceda,~H.~E.; M\"{u}ller,~K.-R. SpookyNet: Learning force fields with electronic degrees of freedom and nonlocal effects. \emph{Nature Communications} \textbf{2021}, \emph{12}\relax
\mciteBstWouldAddEndPuncttrue
\mciteSetBstMidEndSepPunct{\mcitedefaultmidpunct}
{\mcitedefaultendpunct}{\mcitedefaultseppunct}\relax
\EndOfBibitem
\bibitem[Grimme(2006)]{d2prior}
Grimme,~S. Semiempirical GGA‐type density functional constructed with a long‐range dispersion correction. \emph{Journal of Computational Chemistry} \textbf{2006}, \emph{27}, 1787–1799\relax
\mciteBstWouldAddEndPuncttrue
\mciteSetBstMidEndSepPunct{\mcitedefaultmidpunct}
{\mcitedefaultendpunct}{\mcitedefaultseppunct}\relax
\EndOfBibitem
\bibitem[Fey and Lenssen(2019)Fey, and Lenssen]{geometric}
Fey,~M.; Lenssen,~J.~E. Fast Graph Representation Learning with PyTorch Geometric. \emph{CoRR} \textbf{2019}, \emph{abs/1903.02428}\relax
\mciteBstWouldAddEndPuncttrue
\mciteSetBstMidEndSepPunct{\mcitedefaultmidpunct}
{\mcitedefaultendpunct}{\mcitedefaultseppunct}\relax
\EndOfBibitem
\bibitem[Eastman \latin{et~al.}(2023)Eastman, Behara, Dotson, Galvelis, Herr, Horton, Mao, Chodera, Pritchard, Wang, Fabritiis, and Markland]{spice}
Eastman,~P.; Behara,~P.~K.; Dotson,~D.~L.; Galvelis,~R.; Herr,~J.~E.; Horton,~J.~T.; Mao,~Y.; Chodera,~J.~D.; Pritchard,~B.~P.; Wang,~Y. \latin{et~al.}  {SPICE}, A Dataset of Drug-like Molecules and Peptides for Training Machine Learning Potentials. \emph{Scientific Data} \textbf{2023}, \emph{10}\relax
\mciteBstWouldAddEndPuncttrue
\mciteSetBstMidEndSepPunct{\mcitedefaultmidpunct}
{\mcitedefaultendpunct}{\mcitedefaultseppunct}\relax
\EndOfBibitem
\bibitem[Ramakrishnan \latin{et~al.}(2014)Ramakrishnan, Dral, Rupp, and von Lilienfeld]{qm9}
Ramakrishnan,~R.; Dral,~P.~O.; Rupp,~M.; von Lilienfeld,~O.~A. Quantum chemistry structures and properties of 134 kilo molecules. \emph{Scientific Data} \textbf{2014}, \emph{1}\relax
\mciteBstWouldAddEndPuncttrue
\mciteSetBstMidEndSepPunct{\mcitedefaultmidpunct}
{\mcitedefaultendpunct}{\mcitedefaultseppunct}\relax
\EndOfBibitem
\bibitem[Cheng \latin{et~al.}(2019)Cheng, Engel, Behler, Dellago, and Ceriotti]{water}
Cheng,~B.; Engel,~E.~A.; Behler,~J.; Dellago,~C.; Ceriotti,~M. Ab initio thermodynamics of liquid and solid water. \emph{Proceedings of the National Academy of Sciences} \textbf{2019}, \emph{116}, 1110–1115\relax
\mciteBstWouldAddEndPuncttrue
\mciteSetBstMidEndSepPunct{\mcitedefaultmidpunct}
{\mcitedefaultendpunct}{\mcitedefaultseppunct}\relax
\EndOfBibitem
\bibitem[Chmiela \latin{et~al.}(2017)Chmiela, Tkatchenko, Sauceda, Poltavsky, Sch\"{u}tt, and M\"{u}ller]{Chmiela}
Chmiela,~S.; Tkatchenko,~A.; Sauceda,~H.~E.; Poltavsky,~I.; Sch\"{u}tt,~K.~T.; M\"{u}ller,~K.-R. Machine learning of accurate energy-conserving molecular force fields. \emph{Science Advances} \textbf{2017}, \emph{3}\relax
\mciteBstWouldAddEndPuncttrue
\mciteSetBstMidEndSepPunct{\mcitedefaultmidpunct}
{\mcitedefaultendpunct}{\mcitedefaultseppunct}\relax
\EndOfBibitem
\bibitem[Christensen and lilienfeld(2020)Christensen, and lilienfeld]{rmd17}
Christensen,~A.~S.; lilienfeld,~A.~V. {Revised MD17 dataset (rMD17)}. \textbf{2020}, \relax
\mciteBstWouldAddEndPunctfalse
\mciteSetBstMidEndSepPunct{\mcitedefaultmidpunct}
{}{\mcitedefaultseppunct}\relax
\EndOfBibitem
\bibitem[Chmiela \latin{et~al.}(2023)Chmiela, Vassilev-Galindo, Unke, Kabylda, Sauceda, Tkatchenko, and M\"{u}ller]{md22}
Chmiela,~S.; Vassilev-Galindo,~V.; Unke,~O.~T.; Kabylda,~A.; Sauceda,~H.~E.; Tkatchenko,~A.; M\"{u}ller,~K.-R. Accurate global machine learning force fields for molecules with hundreds of atoms. \emph{Science Advances} \textbf{2023}, \emph{9}\relax
\mciteBstWouldAddEndPuncttrue
\mciteSetBstMidEndSepPunct{\mcitedefaultmidpunct}
{\mcitedefaultendpunct}{\mcitedefaultseppunct}\relax
\EndOfBibitem
\bibitem[Smith \latin{et~al.}(2017)Smith, Isayev, and Roitberg]{ani1}
Smith,~J.~S.; Isayev,~O.; Roitberg,~A.~E. {ANI}-1: an extensible neural network potential with {DFT} accuracy at force field computational cost. \emph{Chemical Science} \textbf{2017}, \emph{8}, 3192--3203\relax
\mciteBstWouldAddEndPuncttrue
\mciteSetBstMidEndSepPunct{\mcitedefaultmidpunct}
{\mcitedefaultendpunct}{\mcitedefaultseppunct}\relax
\EndOfBibitem
\bibitem[Smith \latin{et~al.}(2020)Smith, Zubatyuk, Nebgen, Lubbers, Barros, Roitberg, Isayev, and Tretiak]{ani1x1}
Smith,~J.~S.; Zubatyuk,~R.; Nebgen,~B.; Lubbers,~N.; Barros,~K.; Roitberg,~A.~E.; Isayev,~O.; Tretiak,~S. The {ANI}-1ccx and {ANI}-1x data sets, coupled-cluster and density functional theory properties for molecules. \emph{Scientific Data} \textbf{2020}, \emph{7}\relax
\mciteBstWouldAddEndPuncttrue
\mciteSetBstMidEndSepPunct{\mcitedefaultmidpunct}
{\mcitedefaultendpunct}{\mcitedefaultseppunct}\relax
\EndOfBibitem
\bibitem[Devereux \latin{et~al.}(2020)Devereux, Smith, Huddleston, Barros, Zubatyuk, Isayev, and Roitberg]{ani2x}
Devereux,~C.; Smith,~J.~S.; Huddleston,~K.~K.; Barros,~K.; Zubatyuk,~R.; Isayev,~O.; Roitberg,~A.~E. Extending the Applicability of the ANI Deep Learning Molecular Potential to Sulfur and Halogens. \emph{Journal of Chemical Theory and Computation} \textbf{2020}, \emph{16}, 4192--4202, PMID: 32543858\relax
\mciteBstWouldAddEndPuncttrue
\mciteSetBstMidEndSepPunct{\mcitedefaultmidpunct}
{\mcitedefaultendpunct}{\mcitedefaultseppunct}\relax
\EndOfBibitem
\bibitem[Smith \latin{et~al.}(2018)Smith, Nebgen, Lubbers, Isayev, and Roitberg]{ani1x2}
Smith,~J.~S.; Nebgen,~B.; Lubbers,~N.; Isayev,~O.; Roitberg,~A.~E. Less is more: Sampling chemical space with active learning. \emph{The Journal of Chemical Physics} \textbf{2018}, \emph{148}, 241733\relax
\mciteBstWouldAddEndPuncttrue
\mciteSetBstMidEndSepPunct{\mcitedefaultmidpunct}
{\mcitedefaultendpunct}{\mcitedefaultseppunct}\relax
\EndOfBibitem
\bibitem[Jia \latin{et~al.}(2019)Jia, Padon, Thomas, Warszawski, Zaharia, and Aiken]{dlcompiler}
Jia,~Z.; Padon,~O.; Thomas,~J.; Warszawski,~T.; Zaharia,~M.; Aiken,~A. TASO: Optimizing Deep Learning Computation with Automatic Generation of Graph Substitutions. Proceedings of the 27th ACM Symposium on Operating Systems Principles. New York, NY, USA, 2019; p 47–62\relax
\mciteBstWouldAddEndPuncttrue
\mciteSetBstMidEndSepPunct{\mcitedefaultmidpunct}
{\mcitedefaultendpunct}{\mcitedefaultseppunct}\relax
\EndOfBibitem
\bibitem[Tillet \latin{et~al.}(2019)Tillet, Kung, and Cox]{triton}
Tillet,~P.; Kung,~H.~T.; Cox,~D. Triton: An Intermediate Language and Compiler for Tiled Neural Network Computations. Proceedings of the 3rd ACM SIGPLAN International Workshop on Machine Learning and Programming Languages (MAPL 2019). Phoenix, Arizona, United States, 2019\relax
\mciteBstWouldAddEndPuncttrue
\mciteSetBstMidEndSepPunct{\mcitedefaultmidpunct}
{\mcitedefaultendpunct}{\mcitedefaultseppunct}\relax
\EndOfBibitem
\bibitem[Appleyard \latin{et~al.}(2016)Appleyard, Kocisk{\'{y}}, and Blunsom]{fusionRNN}
Appleyard,~J.; Kocisk{\'{y}},~T.; Blunsom,~P. Optimizing Performance of Recurrent Neural Networks on GPUs. \emph{CoRR} \textbf{2016}, \emph{abs/1604.01946}\relax
\mciteBstWouldAddEndPuncttrue
\mciteSetBstMidEndSepPunct{\mcitedefaultmidpunct}
{\mcitedefaultendpunct}{\mcitedefaultseppunct}\relax
\EndOfBibitem
\bibitem[Wang \latin{et~al.}(2010)Wang, Lin, and Yi]{fusion2010}
Wang,~G.; Lin,~Y.; Yi,~W. Kernel fusion: An effective method for better power efficiency on multithreaded GPU. 2010 IEEE/ACM Int'l Conference on Green Computing and Communications \& Int'l Conference on Cyber, Physical and Social Computing. 2010; pp 344--350\relax
\mciteBstWouldAddEndPuncttrue
\mciteSetBstMidEndSepPunct{\mcitedefaultmidpunct}
{\mcitedefaultendpunct}{\mcitedefaultseppunct}\relax
\EndOfBibitem
\bibitem[Filipovi{\v{c}} \latin{et~al.}(2015)Filipovi{\v{c}}, Madzin, Fousek, and Matyska]{fusionblas}
Filipovi{\v{c}},~J.; Madzin,~M.; Fousek,~J.; Matyska,~L. Optimizing CUDA code by kernel fusion: application on BLAS. \emph{The Journal of Supercomputing} \textbf{2015}, \emph{71}, 3934--3957\relax
\mciteBstWouldAddEndPuncttrue
\mciteSetBstMidEndSepPunct{\mcitedefaultmidpunct}
{\mcitedefaultendpunct}{\mcitedefaultseppunct}\relax
\EndOfBibitem
\bibitem[Godwin* \latin{et~al.}(2020)Godwin*, Keck*, Battaglia, Bapst, Kipf, Li, Stachenfeld, Veli\v{c}kovi\'{c}, and Sanchez-Gonzalez]{jraph2020github}
Godwin*,~J.; Keck*,~T.; Battaglia,~P.; Bapst,~V.; Kipf,~T.; Li,~Y.; Stachenfeld,~K.; Veli\v{c}kovi\'{c},~P.; Sanchez-Gonzalez,~A. {J}raph: {A} library for graph neural networks in jax. 2020; \url{http://github.com/deepmind/jraph}\relax
\mciteBstWouldAddEndPuncttrue
\mciteSetBstMidEndSepPunct{\mcitedefaultmidpunct}
{\mcitedefaultendpunct}{\mcitedefaultseppunct}\relax
\EndOfBibitem
\bibitem[Nguyen and Corporation(2008)Nguyen, and Corporation]{celllist}
Nguyen,~H.; Corporation,~N. \emph{GPU Gems 3}; Lab Companion Series v. 3; Addison-Wesley, 2008\relax
\mciteBstWouldAddEndPuncttrue
\mciteSetBstMidEndSepPunct{\mcitedefaultmidpunct}
{\mcitedefaultendpunct}{\mcitedefaultseppunct}\relax
\EndOfBibitem
\bibitem[Tang and Karniadakis(2014)Tang, and Karniadakis]{celllist2}
Tang,~Y.-H.; Karniadakis,~G.~E. Accelerating dissipative particle dynamics simulations on GPUs: Algorithms, numerics and applications. \emph{Computer Physics Communications} \textbf{2014}, \emph{185}, 2809--2822\relax
\mciteBstWouldAddEndPuncttrue
\mciteSetBstMidEndSepPunct{\mcitedefaultmidpunct}
{\mcitedefaultendpunct}{\mcitedefaultseppunct}\relax
\EndOfBibitem
\bibitem[Doerr \latin{et~al.}(2021)Doerr, Majewski, P{\'{e}}rez, Kr\"{a}mer, Clementi, Noe, Giorgino, and Fabritiis]{torchmd}
Doerr,~S.; Majewski,~M.; P{\'{e}}rez,~A.; Kr\"{a}mer,~A.; Clementi,~C.; Noe,~F.; Giorgino,~T.; Fabritiis,~G.~D. {TorchMD}: A Deep Learning Framework for Molecular Simulations. \emph{Journal of Chemical Theory and Computation} \textbf{2021}, \emph{17}, 2355--2363\relax
\mciteBstWouldAddEndPuncttrue
\mciteSetBstMidEndSepPunct{\mcitedefaultmidpunct}
{\mcitedefaultendpunct}{\mcitedefaultseppunct}\relax
\EndOfBibitem
\bibitem[Schoenholz and Cubuk(2020)Schoenholz, and Cubuk]{jaxmd2020}
Schoenholz,~S.~S.; Cubuk,~E.~D. JAX M.D. A Framework for Differentiable Physics. Advances in Neural Information Processing Systems. 2020\relax
\mciteBstWouldAddEndPuncttrue
\mciteSetBstMidEndSepPunct{\mcitedefaultmidpunct}
{\mcitedefaultendpunct}{\mcitedefaultseppunct}\relax
\EndOfBibitem
\bibitem[Larsen \latin{et~al.}(2017)Larsen, Mortensen, Blomqvist, Castelli, Christensen, Dułak, Friis, Groves, Hammer, Hargus, Hermes, Jennings, Jensen, Kermode, Kitchin, Kolsbjerg, Kubal, Kaasbjerg, Lysgaard, Maronsson, Maxson, Olsen, Pastewka, Peterson, Rostgaard, Schiøtz, Schütt, Strange, Thygesen, Vegge, Vilhelmsen, Walter, Zeng, and Jacobsen]{ase-paper}
Larsen,~A.~H.; Mortensen,~J.~J.; Blomqvist,~J.; Castelli,~I.~E.; Christensen,~R.; Dułak,~M.; Friis,~J.; Groves,~M.~N.; Hammer,~B.; Hargus,~C. \latin{et~al.}  The atomic simulation environment—a Python library for working with atoms. \emph{Journal of Physics: Condensed Matter} \textbf{2017}, \emph{29}, 273002\relax
\mciteBstWouldAddEndPuncttrue
\mciteSetBstMidEndSepPunct{\mcitedefaultmidpunct}
{\mcitedefaultendpunct}{\mcitedefaultseppunct}\relax
\EndOfBibitem
\bibitem[pym()]{pymatgen}
{Pymatgen (Python Materials Genomics)}. \url{https://pymatgen.org}, (Accessed April 17 2024)\relax
\mciteBstWouldAddEndPuncttrue
\mciteSetBstMidEndSepPunct{\mcitedefaultmidpunct}
{\mcitedefaultendpunct}{\mcitedefaultseppunct}\relax
\EndOfBibitem
\bibitem[Anderson \latin{et~al.}(2019)Anderson, Hy, and Kondor]{cormorant}
Anderson,~B.; Hy,~T.-S.; Kondor,~R. Cormorant: Covariant Molecular Neural Networks. Proceedings of the 33rd International Conference on Neural Information Processing Systems. Red Hook, NY, USA, 2019\relax
\mciteBstWouldAddEndPuncttrue
\mciteSetBstMidEndSepPunct{\mcitedefaultmidpunct}
{\mcitedefaultendpunct}{\mcitedefaultseppunct}\relax
\EndOfBibitem
\bibitem[Brandstetter \latin{et~al.}(2021)Brandstetter, Hesselink, van~der Pol, Bekkers, and Welling]{segnn}
Brandstetter,~J.; Hesselink,~R.; van~der Pol,~E.; Bekkers,~E.~J.; Welling,~M. Geometric and Physical Quantities Improve E(3) Equivariant Message Passing. arXiv, 2021; 10.48550/ARXIV.2110.02905\relax
\mciteBstWouldAddEndPuncttrue
\mciteSetBstMidEndSepPunct{\mcitedefaultmidpunct}
{\mcitedefaultendpunct}{\mcitedefaultseppunct}\relax
\EndOfBibitem
\bibitem[Sch\"{u}tt \latin{et~al.}(2018)Sch\"{u}tt, Sauceda, Kindermans, Tkatchenko, and M\"{u}ller]{schnet}
Sch\"{u}tt,~K.~T.; Sauceda,~H.~E.; Kindermans,~P.-J.; Tkatchenko,~A.; M\"{u}ller,~K.-R. {SchNet} {\textendash} A deep learning architecture for molecules and materials. \emph{The Journal of Chemical Physics} \textbf{2018}, \emph{148}, 241722\relax
\mciteBstWouldAddEndPuncttrue
\mciteSetBstMidEndSepPunct{\mcitedefaultmidpunct}
{\mcitedefaultendpunct}{\mcitedefaultseppunct}\relax
\EndOfBibitem
\bibitem[Liao and Smidt(2023)Liao, and Smidt]{equiformer}
Liao,~Y.-L.; Smidt,~T. Equiformer: Equivariant Graph Attention Transformer for 3D Atomistic Graphs. The Eleventh International Conference on Learning Representations. 2023\relax
\mciteBstWouldAddEndPuncttrue
\mciteSetBstMidEndSepPunct{\mcitedefaultmidpunct}
{\mcitedefaultendpunct}{\mcitedefaultseppunct}\relax
\EndOfBibitem
\bibitem[Gasteiger \latin{et~al.}(2020)Gasteiger, Giri, Margraf, and G\"{u}nnemann]{dimenet++}
Gasteiger,~J.; Giri,~S.; Margraf,~J.~T.; G\"{u}nnemann,~S. Fast and Uncertainty-Aware Directional Message Passing for Non-Equilibrium Molecules. arXiv, 2020; 10.48550/ARXIV.2011.14115\relax
\mciteBstWouldAddEndPuncttrue
\mciteSetBstMidEndSepPunct{\mcitedefaultmidpunct}
{\mcitedefaultendpunct}{\mcitedefaultseppunct}\relax
\EndOfBibitem
\bibitem[Liu \latin{et~al.}(2021)Liu, Wang, Liu, Zhang, Oztekin, and Ji]{spherenet}
Liu,~Y.; Wang,~L.; Liu,~M.; Zhang,~X.; Oztekin,~B.; Ji,~S. Spherical Message Passing for 3D Graph Networks. arXiv, 2021; 10.48550/ARXIV.2102.05013\relax
\mciteBstWouldAddEndPuncttrue
\mciteSetBstMidEndSepPunct{\mcitedefaultmidpunct}
{\mcitedefaultendpunct}{\mcitedefaultseppunct}\relax
\EndOfBibitem
\bibitem[Fu \latin{et~al.}(2023)Fu, Wu, Wang, Xie, Keten, Gomez-Bombarelli, and Jaakkola]{stable1}
Fu,~X.; Wu,~Z.; Wang,~W.; Xie,~T.; Keten,~S.; Gomez-Bombarelli,~R.; Jaakkola,~T.~S. Forces are not Enough: Benchmark and Critical Evaluation for Machine Learning Force Fields with Molecular Simulations. \emph{Transactions on Machine Learning Research} \textbf{2023}, Survey Certification\relax
\mciteBstWouldAddEndPuncttrue
\mciteSetBstMidEndSepPunct{\mcitedefaultmidpunct}
{\mcitedefaultendpunct}{\mcitedefaultseppunct}\relax
\EndOfBibitem
\bibitem[Vita and Schwalbe-Koda(2023)Vita, and Schwalbe-Koda]{stable2}
Vita,~J.~A.; Schwalbe-Koda,~D. Data efficiency and extrapolation trends in neural network interatomic potentials. \emph{Machine Learning: Science and Technology} \textbf{2023}, \emph{4}, 035031\relax
\mciteBstWouldAddEndPuncttrue
\mciteSetBstMidEndSepPunct{\mcitedefaultmidpunct}
{\mcitedefaultendpunct}{\mcitedefaultseppunct}\relax
\EndOfBibitem
\bibitem[Wagner \latin{et~al.}(2023)Wagner, Thompson, Mobley, Chodera, Bannan, Rizzi, trevorgokey, Dotson, Mitchell, jaimergp, Camila, Behara, Bayly, JoshHorton, Wang, Pulido, Lim, Sasmal, SimonBoothroyd, Dalke, Smith, Horton, Wang, Gowers, Zhao, Davel, and Zhao]{jeff_wagner_2023_10103216}
Wagner,~J.; Thompson,~M.; Mobley,~D.~L.; Chodera,~J.; Bannan,~C.; Rizzi,~A.; trevorgokey; Dotson,~D.~L.; Mitchell,~J.~A.; jaimergp \latin{et~al.}  {openforcefield/openff-toolkit: 0.14.5 Minor feature release}. 2023; \url{https://doi.org/10.5281/zenodo.10103216}\relax
\mciteBstWouldAddEndPuncttrue
\mciteSetBstMidEndSepPunct{\mcitedefaultmidpunct}
{\mcitedefaultendpunct}{\mcitedefaultseppunct}\relax
\EndOfBibitem
\bibitem[Grimme \latin{et~al.}(2017)Grimme, Bannwarth, and Shushkov]{grimme2017robust}
Grimme,~S.; Bannwarth,~C.; Shushkov,~P. A robust and accurate tight-binding quantum chemical method for structures, vibrational frequencies, and noncovalent interactions of large molecular systems parametrized for all spd-block elements (Z= 1--86). \emph{Journal of chemical theory and computation} \textbf{2017}, \emph{13}, 1989--2009\relax
\mciteBstWouldAddEndPuncttrue
\mciteSetBstMidEndSepPunct{\mcitedefaultmidpunct}
{\mcitedefaultendpunct}{\mcitedefaultseppunct}\relax
\EndOfBibitem
\bibitem[Sun \latin{et~al.}(2018)Sun, Berkelbach, Blunt, Booth, Guo, Li, Liu, McClain, Sayfutyarova, Sharma, \latin{et~al.} others]{sun2018pyscf}
Sun,~Q.; Berkelbach,~T.~C.; Blunt,~N.~S.; Booth,~G.~H.; Guo,~S.; Li,~Z.; Liu,~J.; McClain,~J.~D.; Sayfutyarova,~E.~R.; Sharma,~S. \latin{et~al.}  PySCF: the Python-based simulations of chemistry framework. \emph{Wiley Interdisciplinary Reviews: Computational Molecular Science} \textbf{2018}, \emph{8}, e1340\relax
\mciteBstWouldAddEndPuncttrue
\mciteSetBstMidEndSepPunct{\mcitedefaultmidpunct}
{\mcitedefaultendpunct}{\mcitedefaultseppunct}\relax
\EndOfBibitem
\end{mcitethebibliography}
\appendix

\section{Hyperparameters}
In the pursuit of transparency and reproducibility, this appendix provides a detailed account of the hyperparameters employed in our computational experiments. The tables contained herein present the specific values and settings used to achieve the results discussed in the main body of this paper. Readers and fellow researchers are encouraged to refer to these tables when attempting to replicate our results or when utilizing the torchmd-train utility for their own training purposes.

\begin{table}
\centering
\caption{MD17 hyperparameters used for the ET training.}
\label{tab:md17}

\begin{tabular}{lc}
\hline
\textbf{Parameter} & \textbf{Value} \\
\hline
\texttt{activation} & \texttt{silu} \\
\texttt{attn\_activation} & \texttt{silu} \\
\texttt{batch\_size} & \texttt{8} \\
\texttt{cutoff\_lower} & \texttt{0.0} \\
\texttt{cutoff\_upper} & \texttt{5.0} \\
\texttt{derivative} & \texttt{True} \\
\texttt{distance\_influence} & \texttt{both} \\
\texttt{early\_stopping\_patience} & \texttt{300} \\
\texttt{ema\_alpha\_neg\_dy} & \texttt{1.0}\\
\texttt{ema\_alpha\_y} & \texttt{0.05}\\
\texttt{embedding\_dimension} & \texttt{128} \\
\texttt{lr} & \texttt{1e-3} \\
\texttt{lr\_factor} & \texttt{0.8} \\
\texttt{lr\_min} & \texttt{1e-7} \\
\texttt{lr\_patience} & \texttt{30} \\
\texttt{lr\_warmup\_steps} & \texttt{1000} \\
\texttt{neg\_dy\_weight} & \texttt{0.8} \\
\texttt{num\_heads} & \texttt{8}\\
\texttt{num\_layers} & \texttt{6} \\
\texttt{num\_rbf} & \texttt{32} \\
\texttt{seed} & \texttt{1} \\
\texttt{train\_size} & \texttt{950} \\
\texttt{val\_size} & \texttt{50} \\
\texttt{vector\_cutoff} & \texttt{True} \\
\texttt{y\_weight} & \texttt{0.2} \\
\hline
\end{tabular}
\end{table}

\begin{table}
\centering
\caption{QM9 $U_0$ hyperparameters used to obtain the results with ET$_{\mathbf{new}}$.}
\label{tab:qm9et}

\begin{tabular}{lc}
\hline
\textbf{Parameter} & \textbf{Value} \\
\hline
\texttt{activation} & \texttt{silu} \\
\texttt{attn\_activation} & \texttt{silu} \\
\texttt{batch\_size} & \texttt{128} \\
\texttt{cutoff\_lower} & \texttt{0.0} \\
\texttt{cutoff\_upper} & \texttt{5.0} \\
\texttt{derivative} & \texttt{False} \\
\texttt{distance\_influence} & \texttt{both} \\
\texttt{early\_stopping\_patience} & \texttt{150} \\
\texttt{ema\_alpha\_neg\_dy} & \texttt{1.0}\\
\texttt{ema\_alpha\_y} & \texttt{1.0}\\
\texttt{embedding\_dimension} & \texttt{256} \\
\texttt{lr} & \texttt{4e-4} \\
\texttt{lr\_factor} & \texttt{0.8} \\
\texttt{lr\_min} & \texttt{1e-7} \\
\texttt{lr\_patience} & \texttt{15} \\
\texttt{lr\_warmup\_steps} & \texttt{10000} \\
\texttt{neg\_dy\_weight} & \texttt{0.0} \\
\texttt{num\_heads} & \texttt{8}\\
\texttt{num\_layers} & \texttt{8} \\
\texttt{num\_rbf} & \texttt{64} \\
\texttt{remove\_ref\_energy} & \texttt{true} \\
\texttt{seed} & \texttt{1} \\
\texttt{train\_size} & \texttt{110000} \\
\texttt{val\_size} & \texttt{10000} \\
\texttt{vector\_cutoff} & \texttt{True} \\
\texttt{y\_weight} & \texttt{1.0} \\
\hline
\end{tabular}
\end{table}

\begin{table}
\centering
\caption{QM9 $U_0$ hyperparameters used for trainings with TensorNet 3L*.}
\label{tab:qm9}
\begin{tabular}{lc}
\hline
\textbf{Parameter} & \textbf{Value} \\
\hline
\texttt{activation} & \texttt{silu} \\
\texttt{batch\_size} & \texttt{16} \\
\texttt{cutoff\_lower} & \texttt{0.0} \\
\texttt{cutoff\_upper} & \texttt{5.0} \\
\texttt{derivative} & \texttt{False} \\
\texttt{early\_stopping\_patience} & \texttt{150} \\
\texttt{embedding\_dimension} & \texttt{256} \\
\texttt{equivariance\_invariance\_group} & \texttt{O(3)} \\
\texttt{gradient\_clipping} & \texttt{40} \\
\texttt{lr} & \texttt{1e-4} \\
\texttt{lr\_factor} & \texttt{0.8} \\
\texttt{lr\_min} & \texttt{1e-7} \\
\texttt{lr\_patience} & \texttt{15} \\
\texttt{lr\_warmup\_steps} & \texttt{1000} \\
\texttt{neg\_dy\_weight} & \texttt{0.0} \\
\texttt{num\_layers} & \texttt{3} \\
\texttt{num\_rbf} & \texttt{64} \\
\texttt{remove\_ref\_energy} & \texttt{true} \\
\texttt{seed} & \texttt{2} \\
\texttt{train\_size} & \texttt{110000} \\
\texttt{val\_size} & \texttt{10000} \\
\texttt{y\_weight} & \texttt{1.0} \\
\hline

\end{tabular}
\end{table}

\begin{table}
\centering
\caption{ANI2x hyperparameters used for training TensorNet for use to run the stable molecular dynamics simulations.}
\label{tab:TN2L_ANI2x}

\begin{tabular}{lc}
\hline
\textbf{Parameter} & \textbf{Value} \\
\hline
\texttt{activation} & \texttt{silu} \\
\texttt{batch\_size} & \texttt{256} \\
\texttt{cutoff\_lower} & \texttt{0.0} \\
\texttt{cutoff\_upper} & \texttt{5.0} \\
\texttt{derivative} & \texttt{True} \\
\texttt{early\_stopping\_patience} & \texttt{50} \\
\texttt{embedding\_dimension} & \texttt{128} \\
\texttt{equivariance\_invariance\_group} & \texttt{O(3)} \\
\texttt{gradient\_clipping} & \texttt{100} \\
\texttt{lr} & \texttt{1e-3} \\
\texttt{lr\_factor} & \texttt{0.5} \\
\texttt{lr\_min} & \texttt{1e-7} \\
\texttt{lr\_patience} & \texttt{4} \\
\texttt{lr\_warmup\_steps} & \texttt{1000} \\
\texttt{neg\_dy\_weight} & \texttt{100} \\
\texttt{num\_layers} & \texttt{\{0},\texttt{2\}} \\
\texttt{num\_rbf} & \texttt{32} \\
\texttt{seed} & \texttt{1} \\
\texttt{train\_size} & \texttt{0.9} \\
\texttt{val\_size} & \texttt{0.1} \\
\texttt{y\_weight} & \texttt{1.0} \\
\hline
\end{tabular}
\end{table}
\end{document}